%% file: main.tex
\documentclass[sigconf]{acmart}
\AtBeginDocument{%
  \providecommand\BibTeX{{%
    \normalfont B\kern-0.5em{\scshape i\kern-0.25em b}\kern-0.8em\TeX}}}

\usepackage{times}
\usepackage{soul}
\usepackage{url}

\usepackage[font=small]{caption}
\usepackage{graphicx}
\usepackage{subcaption}
\usepackage{amsmath,amsfonts} 
\usepackage[ruled,linesnumbered]{algorithm2e}
\usepackage{amsthm}
\usepackage{amsmath}
{
      \theoremstyle{plain}
      \newtheorem{assumption}{Assumption}
      \newtheorem{definition}{Definition}
  }
\usepackage{multirow}
\usepackage{booktabs}
\usepackage{algorithmic}
\usepackage{natbib}
\usepackage[misc]{ifsym}


\begin{document}

\title[Knowledge Guided Representation Learning and Causal Structure Learning in Soil Science]{Knowledge Guided Representation Learning\\ and Causal Structure Learning in Soil Science}

\author{Somya Sharma}
\affiliation{%
  \institution{University of Minnesota}
  \country{USA}
}
\author{Swati Sharma}
\affiliation{%
  \institution{Microsoft Research}
   \city{Redmond}
   \country{USA}}
\email{swatisharma@microsoft.org}
\author{Licheng Liu}
\affiliation{%
 \institution{University of Minnesota}
 \country{USA}}
\author{Rishabh Tushir}
\affiliation{%
  \institution{University of Glasgow}
  \country{UK}}
  \author{Andy Neal}
\affiliation{%
  \institution{Rothamsted Research}
  \country{UK}}
\author{Robert Ness}
\affiliation{%
  \institution{Microsoft Research}
   \city{Redmond}
  \country{USA}}
\author{Vipin Kumar}
\affiliation{%
  \institution{University of Minnesota}
  \country{USA}
  }
\author{John Crawford}
\affiliation{%
  \institution{University of Glasgow}
  \country{UK}}
\author{Emre Kiciman}
\affiliation{%
  \institution{Microsoft Research}
     \city{Redmond}
  \country{USA}}
\author{Ranveer Chandra}
\affiliation{%
  \institution{Microsoft Research}
   \city{Redmond}
   \country{USA}}




\renewcommand{\shortauthors}{Sharma and Sharma, et al.}

\begin{abstract}
     

    An improved understanding of soil can enable more sustainable land-use practices. Nevertheless, soil is called a complex, living medium due to the complex interaction of different soil processes that limit our understanding of soil. Process-based models and analyzing observed data provide two avenues for improving our understanding of soil processes. Collecting observed data is cost-prohibitive but reflects real-world behavior, while process-based models can be used to generate ample synthetic data which may not be representative of reality. We propose a framework, knowledge-guided representation learning, and causal structure learning (KGRCL), to accelerate scientific discoveries in soil science. The framework improves representation learning for simulated soil processes via conditional distribution matching with observed soil processes. Simultaneously, the framework leverages both observed and simulated data to learn a causal structure among the soil processes. The learned causal graph is more representative of ground truth than other graphs generated from other causal discovery methods. Furthermore, the learned causal graph is leveraged in a supervised learning setup to predict the impact of fertilizer use and changing weather on soil carbon. We present the results in five different locations to show the improvement in the prediction performance in out-of-sample and few-shots setting. 
    
\end{abstract}

\maketitle

\input{introduction.tex}
\input{prob_def.tex}

\input{method.tex}

\section{Experiment Results and Discussion}
 \input{data.tex}

\input{results.tex}

\input{discussion.tex}

\input{related_work.tex}

\input{conclusion.tex}
\bibliographystyle{ACM-Reference-Format}
\bibliography{bibliography} 
\clearpage
\appendix
\input{appendix.tex}




\end{document}

%% file: introduction.tex
\section{Introduction}
Machine learning (ML) models have found tremendous success in commercial applications. Recently, these methods have increasingly been adopted by domain experts in scientific communities for improving understanding of real-world processes~\cite{dlphysics}. Although ML methods have demonstrated an improved ability in prediction tasks ~\cite{nguyen2021predicting, padarian2020machine}, the reliance of conventional ML on the i.i.d. assumption that training data represents the deployed environment, limits their out-of-distribution generalizability~\cite{ wadoux2020machine, padarian2020machine,  grunwald2022artificial, emadi2020predicting, ren2022dice}. To improve this, either large-scale, diverse data sets are utilized or careful ML model architectural modifications are needed to emulate the behavior of real-world physical systems. In these scenarios, architectural modifications are themselves limited by domain experts' understanding of the physical systems~\cite{lavin2021simulation}. Causal approaches such as causal discovery and causal effect inference are effective tools that help overcome the limitations of ML approaches~\cite{pearl2018}. These approaches utilize or estimate the cause-and-effect relations among the physical processes explicitly and help provide robust predictions.

In this paper, 
we propose a causal structure learning method, Knowledge-Guided Representation learning and Causal Learning (KGRCL), that jointly learns causal structures from both simulated and observed data. Both sets of data provide heterogeneous sources of information about the same underlying processes. This joint modeling allows us to learn richer representations of complex systems by (a) incorporating observed data to mitigate bias introduced by the simplifying modeling assumptions of simulators, and (b) utilizing simulated data to overcome data sparsity commonly encountered in observed datasets. Simulated data also offers avenues to learn system characteristics that may be unobserved or difficult to observe. 
To reduce the bias introduced by the simulator's data generating process, we use conditional distribution matching~\cite{guo2021outofdistribution} to estimate simulated data distributions that are closer to real world data.

We demonstrate KGCRL in a case study of soil organic carbon modeling using data from farms located across the UK and the USA. Our aim is to understand the factors that cause changes in soil organic carbon. Soil organic carbon (SOC) is the carbon component of organic compounds found in soil both as biomass and as sequestered compounds and necromass. Studying  SOC is of great significance because it is considered to be a ``natural insurance against climate change'' \cite{droste2020soil}---with evidence associating increased soil organic matter with increased crop yields~\cite{SCHJONNING201835, soil-5-15-2019, Lal27A}. Proper management of soil, including its organic carbon component, can mitigate shortages in food, water, energy and adverse repercussions of climate change~\cite{de2016towards}. 
Measuring and monitoring soil organic carbon can therefore have a positive impact in solving several environmental problems. This has led to increased interest by environmentalists, economists and soil scientists, as interdisciplinary collaborations, in improving public awareness and policy making ~\cite{lal2004managing, nziguheba2015soil, white2005principles, bhattacharyya2015soil, minasny2017soil, lal2004soil}.

While the problem of studying soil organic carbon is well-motivated, forming hypotheses and designing experiments to estimate soil organic carbon can be challenging. Changes in soil organic carbon are not only dictated by weather events and management practices, but also by other soil processes such as plant nutrient uptake, soil organisms, soil texture, micro-nutrient content and soil disturbance. This makes soil a complex ``living'' porous medium \cite{de2014soil, de2016towards, ohlson2014soil}. 
Due to the complex nature of soil science, the exact relations among all soil processes is not yet known. There is no accepted universal method for studying soil organic carbon and the relation among soil processes \cite{bouma2002land}. Moreover, current models of soil organic carbon (RothC-26.3~\cite{rothc}, Century~\cite{century}, DNDC~\cite{GILTRAP2010292}, 
are not consistent with the latest advances in understanding of soil processes and do not generalize to different soil conditions found globally as they are limited by their modeling assumptions\cite{modelingsoilprocesses}. 
Therefore, generalizable ML methods can be pivotal in standardizing efforts for soil carbon measuring, reporting and verification across different regions, soil conditions and land use types. Therefore, in our paper, we also investigate if our KGRCL causal structure learning method can help improve our understanding of soil organic carbon. We summarize our key contributions below:
\begin{itemize}
    \item We propose a deep causal structure learning approach that aims to match representations of simulated data to real-world data using conditonal distribution matching. Our causal structure learning experiments in section ~\ref{sec:res_causal_dis} show that KGRCL outperforms other causal discovery methods. 
    \item We use the learned causal structure in zero-shot and few-shot out-of-distribution downstream prediction tasks. Our results indicate that integrating the underlying causal structure in our supervised learning model results in lower errors compared to conventional ML approaches. In addition, the performance of the proposed approach surpasses that of other causally informed approaches (section 
    ~\ref{sec:res_causal_gnn}). 
    \item In this work, we bring together an interdisciplinary team of computer scientists and soil experts to help formulate and validate the framework for soil organic carbon modeling. We evaluate the effectiveness of the proposed algorithm empirically on soil data sets from multiple locations in the UK and the USA. For these experiments, we include widely-used process models for simulating soil organic carbon and related soil processes~\cite{li1996dndc, gmd-15-2839-2022}. We briefly talk about the impact of this work for farmers and soil experts in section~\ref{sec:discussion}.     
\end{itemize}

%% file: prob_def.tex
\section{Preliminaries}
Our goal in this work is twofold. First we learn a causal graph $G$ from multiple data sources, $X$. Next, we use the estimated graph $G$ for out-of-distribution downstream prediction tasks. 

Let $X=\{X_{obs}, X_{sim}\}$ represent the covariates derived from observed ($X_{obs}$) and simulated ($X_{sim}$) data, where observed data $x_{obs}$ and simulated data $x_{sim}$ are drawn from data distribution $q_{obs}$ and $q_{sim}$ such that $X_{obs}\subseteq \mathbb{R}^{p}$ and $X_{sim}\subseteq \mathbb{R}^{d}$. $p$ and $d$ are the number of observed and simulated features. We assume the existence of common features among the two data sets. Let $c$ be the number of common features. 
We can define the causal graph as follows~\cite{pearl2000models, spirtes2000causation}. 
\begin{definition}{(Causal Graph)}
Causal graphs are directed acyclic graphs (DAGs) that encode assumptions about the data generating process. Let $G$ be a DAG where the nodes and the edges represent features in $X$ and, the relationships between them, respectively. Therefore, $G = (V, E)$ consists of vertex set $V:= \{1,2,...(p+d-c)\}$, in which each node $i$ corresponds with variable $X_i$ and an edge set $E \subseteq V^2$, in which, edge $(i,j) \in E$ if $X_i \in Pa_{X_j}$.  
\end{definition}
\noindent where, $Pa_{X_j}$ refers to the parents of $X_j$. 
We further make the following assumptions that enable the estimated graph to have causal interpretation. 
\begin{assumption}{(Faithfulness)}
We assume that the joint distribution $\mathcal{P}_X$ and DAG $G$ are faithful to each other. Therefore, $G$ can be recovered from the data samples $X$ using structure learning.
\end{assumption}

\begin{assumption}{(Causal Sufficiency)}
We assume that there are no latent causes of the observed variables. Therefore, due to the acyclicity, the DAG, $G$, satisfies the Markov assumption.   
\end{assumption}

\noindent In our empirical analysis, we attempt to satisfy the causal sufficiency assumption by including covariates from diverse and rich data sets. These data sets not only include observed variables, but also have covariates obtained from process-based models. These covariates reflect the in-depth and extensive physical knowledge encapsulated in these process-based models. Furthermore, to learn an acyclical graph, we include a constraint to enforce the assumption in the loss function, as given below. 

\begin{theorem}[Acyclicity \cite{DAGGNN}]
    Let $G \in \mathbb{R}^{(p+d-c)}$ be the adjacency matrix used to represent the graph. For an $\alpha > 0$, the graph is acyclic if and only if,

    \begin{equation*}
        tr[(I + \alpha G \odot G)^m] - m = 0
    \end{equation*}
\end{theorem}

\noindent We further assume that there exists a parameter set $\phi$ such that $p_{\phi} (X;G^0) = p(X, G^0)$, where $G^0$ represents the true causal DAG. Similar to the analysis in ~\cite{geffner2022deep}, the universal approximation capabilities of our neural network based structure learning model allows the model to be highly flexible. This increases the likelihood of our learned graph to be correctly specified. 

 To facilitate learning from multiple data sources, conditional distribution matching methods provide a pathway for domain adaptation and have been widely employed to handle distributional shifts across data sources~\cite{guo2021outofdistribution}. 
 We incorporate conditional distribution matching in our framework as,
 \begin{equation}
     \arg \min_{\phi} \hspace{3pt} KL (p(\mathcal{F}(X)|E = e)|| p(\mathcal{F}(X)| E = e')).
 \end{equation}
\noindent In our formulation, the domain or data sets $E$ can take values $e=\text{simulated}$ and $e'=\text{observed}$, and $\mathcal{F}$ is a neural network that learns the representations for the data sets as part of the VAE model.

Further, the learned $G$ can be used in supervised learning tasks. In our experiments $G$ is used to define graph skeletons in graph neural networks to incorporate inductive biases about soil processes in different farmlands. For prediction tasks in out-of-sample scenarios, where $p_{train}(x, y) \neq p_{test}(x, y)$. Our goal is to estimate $\mathbb{E}_{test}(Y|x)$ given only $p_{train}(x, y)$. For this class of problems, causal methods appear intuitively because of the principal of invariance~\cite{pearl_2009}, which assumes that the conditional distribution of $Y$ given its direct causes does not change when intervening on variables other than $Y$. It is shown in~\cite{rojas2018} that such a causal regression model is optimal in the robust sense. For this reason, causal models have been shown to be useful in out-of-distribution scenarios ~\cite{pearltransportability, causalanticausallearning, pmlr-v28-zhang13d, zhangcausal, ChristiansesCausalReg}. 
We assume that all relevant causal variables are observed or present in one or both data modalities. 
Also, we assume that either there is common support $supp_{train}(pa(Y))=supp_{test}(pa(Y))$ or extrapolation is possible from $supp_{train}(pa(Y))$ to $supp_{test}(pa(Y))$, where $pa(Y)$ are the direct causes of $Y$. 
In summary, for the regression task, instead of measuring the conditional expected response $\mathbb{E}(Y|X)$,  we evaluate $\mathbb{E}(Y|X, G)$  which is influenced by not only $P(X)$ but also by the causal graph $G$.

%% file: method.tex
\begin{figure*}[t!]

    \centering
    \includegraphics[width=0.8\textwidth]{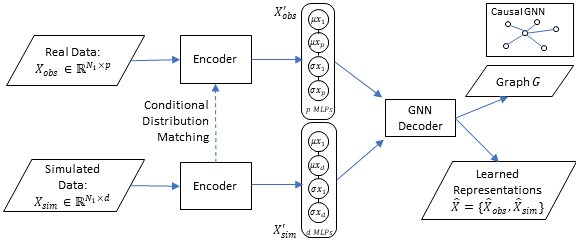}
    \caption{Knowledge-Guided Representation and Causal Learning (KGRCL) framework. (a) Representation learning is facilitated by two encoders that learn hidden representations as part of the GNN based VAE architecture. (b) The simulated data is improved by providing feedback from the hidden encodings learned on the real data. (c) The GNN based decoder is able to infer a causal graph between the variables. (d) As a downstream task, we use the causal graph as a skeleton in GNN model for improved OOD predictions.}
 \label{fig:infographic}   
\end{figure*}

\section{Knowledge-guided Representation Learning and Causal Learning (KGRCL)}


\subsection{GNN based Autoencoder Framework}

Our framework learns from two heterogeneous sources of data, observed, $X_{obs}$, and simulated data, $X_{sim}$. This mutual sharing of information enriches our causal graph, $G$. Figure~\ref{fig:infographic} showcases the KGRCL framework. We propose a GNN based autoencoder framework that learns representations for the covariates $X$ and also learns a causal DAG $G$ among the covariates. Latent representations of real $X_{obs}$ and simulated $X_{sim}$ data are learned using 
the encoder networks. Distributional parameters $\mu$ and $\sigma$ are learned for each of the $p$ features in $X_{obs}$. Similarly, for simulated data $X_{sim}$, distributional parameters $\mu$ and $\sigma$ are learned for each of the $d$ features in $X_{sim}$. We pass this information into the decoder network comprising of a graph neural network with learnable graph $G$. The graph neural network enables us to learn the relation among the $p+d-c$ variables based on the distribution of related hidden encodings, $X' = \{X_{obs}', X_{sim}'\}$. In message passing in the GNN based decoder, the node embeddings are updated in $T$ message passing ( or node-to-message / n2m) and node encoding update (or message-to-node / m2n) operations. These operations at $t^{th}$ iteration are given as,
\begin{equation}
    \text{n2m} : \mathbf{h}_{i \rightarrow j}^{(t), f} = MLP^f ( x_{obs,i}^{'(t-1)}, x_{obs,j}^{'(t-1)}, x_{sim,i}^{(t-1)}, x_{sim,j}^{(t-1)}  ),
\end{equation}
and
\begin{equation}
    \text{m2n} : x_i^{'(t)} = MLP^{m2n} ( \sum_{q \ne i} G_{qi} . \mathbf{h}^{(t), f}_{q \rightarrow i} ).
\end{equation}
In the node-to-message (n2m) update, we aggregate the message from neighboring nodes. A message going from node $i$ to node $j$ at the $t^{th}$ message passing iteration, can be transformed using the non-linear mapping provided by $MLP^f$. The hidden encoding $X'$ at $t^{th}$ iteration is obtained in the m2n operation, where $MLP^{m2n}$ provides an update using the graph representation $G$ and hidden node representation $\mathbf{h}$. The message aggregation relies on hidden embeddings $X'$ that we get from the two sources of data. The updated hidden representations $X'$ is used to create the final feature vector $\hat{X}$ through the read-out MLP layer. In each epoch, we update our representations and causal graph. Therefore, for $k^{th}$ sample, equations \ref{eq:n2m} and \ref{eq:m2n} can be written as,

\begin{equation}
    \text{n2m}_k : \mathbf{h}_{i_k \rightarrow j_k}^{(t), f} = MLP^f ( x_{obs,i_k}^{'(t-1)}, x_{obs,j_k}^{'(t-1)}, x_{sim,i_k}^{(t-1)}, x_{sim,j_k}^{(t-1)}  ),
    \label{eq:n2m}
\end{equation}

and

\begin{equation}
    \text{m2n}_k : x_{i_k}^{'(t)} = MLP^{m2n} ( \sum_{q \ne i} G_{q_k i_k} . \mathbf{h}^{(t), f}_{q_k \rightarrow i_k} ),
    \label{eq:m2n}
\end{equation}




Similar to other message passing neural networks \cite{gilmer2017neural}, the readout MLP, R, helps in obtaining representations for $X$ as, $\hat{X} = R(\{ x_i^T | i \in G \})$.



\subsection{Loss Function}

We aim to learn a DAG that fits well to our data $\mathrm{X}$,

\begin{equation}
    \text{arg max}_{G} p(\mathrm{X}|G, X') p(G)
\end{equation}

An explicit regularization term is added to incorporate to enforce the acyclicity constraint on $G$,

\begin{equation}
    \text{arg max}_{G } (p(\mathrm{X}|G, X') p(G|X') + \lambda_A \text{Loss}_A
\end{equation}

In our Bayesian approach, the joint likelihood is,

\begin{equation}
    p(X, X', G) = p(G) \Pi_i p(x_i, X_i', G) p(X_i')
\end{equation}

Due to the intractability of the joint likelihood, we use variational inference~\cite{kingma2013auto}. Similar to~\cite{visl2022}, we consider factorized variational distribution $q(X',G) = q_{\phi}(G) \Pi_i q_{\phi} (X_i'|x_i)$. In our framework, encoders estimate the mean and covariance matrices $q_{\phi}(X_i'|x_i)$.

\noindent \textbf{ELBO} We use separate encoders to estimate the variational parameters for the two data sets. Hence, we can decompose the ELBO cost into separate likelihood and complexity costs for the two data sets. The complexity cost associated with the shared GNN decoder can be computed as a KL divergence term.

The ELBO objective is of the form:
\begin{equation} \label{eq:elbo}
\small
\begin{split}
    \text{ELBO} = &  \sum_n \{  \mathbb{E}_{q_{\phi} (X_n'|X_n) q(G)} [log p(X_n|X_n', G)]   \\
    - & KL (q_{\phi}(X_n'|X_n)||p(X_n')) ] \}\\
    -&  KL (q(G)||p(G))\\
     \\
     = &  \sum_n \{  \mathbb{E}_{q_{\phi_d} (X_{sim,n}'|X_{sim,n}) q(G)} [log p(X_{sim,n}|X_{sim,n}', G)]    \\
    - & KL (q_{\phi_d}(X_{sim,n}'|X_{sim,n})||p(X_{sim,n})) ]  \\ 
    + & \mathbb{E}_{q_{\phi_p} (X'_{obs,n}|X_{obs,n}) q(G)} [log p(X_{obs,n}|X'_{obs,n}, G)]   \\ 
    - &  KL (q_{\phi_p}(X'_{obs,n}|X_{obs,n})||p(X'_{obs,n})) ]  \} \\
     - &   KL (q(G)||p(G))\\
\end{split}
\end{equation}

There are three major terms in the ELBO formulation which are further decomposed based on source of data. Equation~\ref{eq:elbo} looks at the decomposed ELBO formulation with five terms. The first term ensures that $G$ and $X'$ are learned such that likelihood of observing the simulated data is maximized in the objective. This is the likelihood cost on the simulated data while the third term is the likelihood cost on the real data. The second term ensures that we learn a posterior density distribution for our latent representation of simulated data such that it is similar to the prior defined for the latent representation $X'$. Similarly, the fourth term is the complexity cost for the real data ensuring the learned representations are parsimonious like the prior defined for the real data set. The fifth term ensures that the posterior distribution learned for our graph $G$ is similar to the prior defined for it. In this framework, $p(X')$ is a Gaussian prior and $p(G)$ is a Bernoulli prior. The posterior $q(G)$ provides us with the adjacency matrix where each element $p_{ij}$ represents the estimated edge existence probability of an edge going from row $i$ to  column $j$ in the matrix. The initial prior probability values are initialized at 0.5.

\noindent \textbf{Loss}$_{\text{\textbf{DM}}}$\textbf{:} To improve the learned representations on the simulated data, we introduce the conditional distribution matching loss term in Equation~\ref{eq:dm}. For the common or overlapping variables between the simulated and the real data, we can provide feedback to the learned representations of the simulated data from the real data. Equation~\ref{eq:dm} shows that using the learned posterior on the real data as reference, we minimize the KL divergence between the simulated data posterior and the real data posterior for the overlapping variables,
\begin{equation} 
    \text{Loss}_{\text{DM}} = \sum_n \sum_{j\in O} KL ( p (X_{sim, n,j}'|X_{sim, n,j})   || p(X_{obs, n,j}'|X_{obs, n,j})  ).
\label{eq:dm}    
\end{equation}

\noindent Here, $O \subseteq \mathbb{R}^c$ is the set of variables that are overlapping between $X_{obs}$ and $X_{sim}$.

\noindent \textbf{Loss}$_{\text{\textbf{SP}}}$\textbf{:}
For a given downstream task, such as soil carbon prediction, we can use the supervision loss to penalize those reconstructions more that lead to higher error in the predictions of $X_i$,
\begin{equation}
\begin{split}
    \text{Loss}_{\text{SP}} = &  \sum_n \{  \mathbb{E}_{q_{\phi} (X_{obs_i,n}|X'_{obs_i,n}) q(G)} [log p(X_{obs_i,n}|X'_{obs_i, n}, G)].    \\
\end{split}
\end{equation}

\noindent \textbf{Loss}$_{\text{\textbf{A}}}$\textbf{:} Similar to the DAG-GNN~\cite{DAGGNN} framework, we can enable a soft constraint on acyclicity of the graph,

\begin{equation}
     \text{Loss}_{\text{A}} \equiv tr[(I + \alpha G \odot G)^m] - m = 0. 
\end{equation}

The final loss for KGRCL can be represented as,

\begin{equation}
    \mathcal{L} = \text{ELBO} + \lambda_{DM} \text{Loss}_{\text{DM}} + \lambda_{SP} \text{Loss}_{\text{SP}}  + \lambda_{A} 
    \text{Loss}_{\text{A}}.
\label{eq:loss}
\end{equation}

The loss term $\mathcal{L}$ in equation \ref{eq:loss} enables us to learn an improved causal graph structure on the observed data and better representations on the simulated data simultaneously. Since the real world data may contain missing values we can also ensure that we are learning more on the training examples with the observed data. Therefore, we can also introduce a mask based on which values are not missing. This is discussed further in the ablation study in the Results section. KGRCL is summarized in algorithm~\ref{alg:kgrcl}.


\begin{algorithm}
\caption{Knowledge-Guided Representation learning and Causal structure Learning}\label{alg:kgrcl}
\KwIn{Simulated Data $X_{sim} \in \mathbb{R}^{N \times d}$, Observed Data $X_{obs} \in \mathbb{R}^{N \times p}$}

\KwOut{Learned representations $X'_{obs}, X'_{sim}$ and causal graph $G$}

\For {batch $x_{obs}$ and $x_{sim}$ }
{
    Encode $x_{obs}$ and $x_{sim}$ such that $x'_{obs} \sim N(\mu_{obs}, \sigma_{obs}^2)$ and $x'_{sim} \sim N(\mu_{sim}, \sigma_{sim}^2)$
    
    Sample $G$ from $p(G)$ and decode $\hat{x}_{obs}, \hat{x}_{sim} = R(x'_{obs}, x'_{sim}, G)$
    
    Compute the loss as Eq \ref{eq:loss}
    
    Compute gradients and update parameters in encoders, decoder and graph $G$     
}
\end{algorithm}

%% file: data.tex
\subsection{Data}

In this section, we describe the 3 observed and 2 simulated data sets we use in our evaluation framework.

\noindent \textbf{Observed Dataset North Wyke (NW):} 
One of the datasets included in our experiments is North Wyke Farm Platform data, a publicly-available dataset including soil nutrition information for multiple land management practices\footnote{Data available via \url{http://www.rothamsted.ac.uk/north-wyke-farm-platform}}. This data is available for multiple fields located in Devon, UK with three different land use types. 
High resolution long term data including data on soil organic carbon, soil total nitrogen, pH, soil chemistry data as well as management practices are collected to study the sustainability of different land management practices (treatments) over time (2012 to present). We use data from the fields Golden Rove and Great Field, where we have soil organic carbon observations. 


\noindent \textbf{Observed Dataset Washington (WA):} The second dataset is collected on two fields (namely SHOP and Homeplace) situated in Farmington, Washington,  USA. Soil survey data including carbon, nitrogen and pH, field carbon dioxide and methane emissions data, soil chemistry data and details of management practices such as surface tilling, fertilizer, manure application and crop type is available from 2019 to 2021. Weather data for these years were obtained from the weather station data from NOAA. 

\noindent \textbf{Observed dataset Minnesota (MN):}
Another publicly-available dataset that is included in our experiments is the soil sample and emissions data collected in Minnesota. This dataset contains emissions flux and corresponding environment observations during 2016-2018 growing seasons (i.e., April 1st to July 31st) from six chambers in a controlled-environment mesocosm facility at the University of Minnesota. Soil samples were sourced from a farm in Goodhue County, MN, which had been under corn-soybean rotation for 25 years. The observed variables include N\textsubscript{2}O flux, CO\textsubscript{2} flux, air temperature, precipitation, radiation, N fertilizer application, crop planting information, and soil properties including soil NO\textsubscript{3}\textsuperscript{-} concentration, NH\textsubscript{4}\textsuperscript{+} concentration, water content, and soil organic carbon. The details of the mesocosm experiments can be found in ~\cite{MillerEcosys} and the preprocessing setup is similar to ~\cite{gmd-15-2839-2022}.

\noindent \textbf{Simulator DNDC:}
DNDC is a computer-based simulation model for studying the biogeochemistry of carbon (C) and nitrogen (N) in agricultural environments ~\cite{li1996dndc}. The DNDC model has been employed to provide predictions regarding crop output, carbon sequestration, C and N emissions in agroecosystems, and nitrate loss due to leaching. The DNDC model relies on the biogeochemical processes that are frequently mediated by microbes in terrestrial soils. The inputs to DNDC are weather data, soil data including the initial soil organic carbon, bulk density, soil texture, pH, ploughing events and fertilizer inputs. The outputs of the DNDC model are daily soil carbon, nitrogen, water, and climate information. Simulations for North Wyke (3 fields from 2012 to 2021) and Washington (2 fields from 2019 to 2022) datasets are obtained using DNDC. 

\noindent \textbf{Simulator \textit{ecosys}:}
The \textit{ecosys} model is an advanced agroecosystem model constructed from detailed biophysical and biogeochemical rules instead of using empirical relations~\cite{gmd-15-2839-2022}. It represents the evolution in the microbe-associated processes of nitrification and denitrification using substrate kinetics that are sensitive to soil nitrogen availability, soil temperature, soil moisture, and soil oxygen status~\cite{GrantPattyEcosys}. The \textit{ecosys} model was used to generate the synthetic data for the Minnesota site (six experiments from 2016-2018). The input data used to derive these simulations include hourly meteorological (NLDAS-2), layer-wise soil properties (gSSURGO database), and the same crop/fertilizer setup as the Minnesota site. The model outputs include (N\textsubscript{2}O) flux and 76 selected intermediate variables (e.g. CO\textsubscript{2} flux, layer-wise soil water content, and soil nitrogen concentration), covering all variables observed in the Minnesota site.

\begin{figure}
\subfloat{{\includegraphics[width=0.9\linewidth]{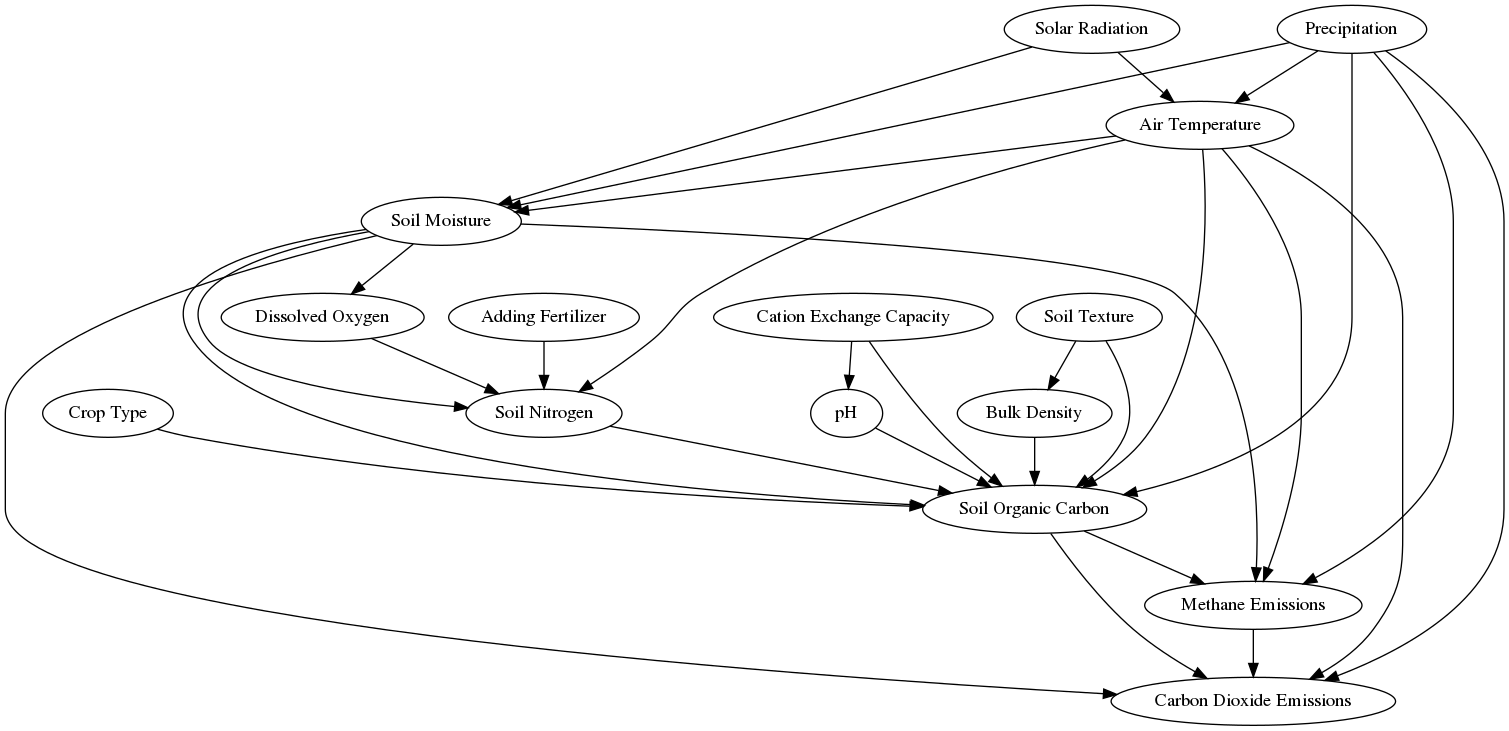} }}%
    \caption{Ground truth causal graph}
    \label{fig:gt}%
\end{figure}

\noindent \textbf{Ground truth for causal graph evaluation:}
Currently, there does not exist a ground truth to validate the results of causal discovery for the three observed datasets. We create a ground truth causal graph (Figure~\ref{fig:gt}) by collecting related research papers and providing them as prompts to GPT-3~\cite{gpt3} to retrieve pairwise relationships for a subset of variables. As an additional step, we got this graph reviewed and validated by soil experts.  

\noindent \paragraph{\textbf{Data preprocessing}} Since the observed datasets have several missing values, we remove variables that contain more than fifty percent missing values. For other variables, we impute the values using a feed-forward neural network that is used to learn representations for the observed dataset based on variables that have complete information. We standardize data using min-max scaling. We resample all the available features to a daily cadence for all the observed datasets. This also enables comparison of the simulated and the observed data at the daily level. 

\subsection{Experimental Setup}

In our setup, we focus on solving two related problems - causal structure learning and soil carbon prediction.

We compare the KGRCL framework with other popular causal structure learning methods. In the loss function $\mathcal{L}$, we fine-tune the penalty coefficients using a grid search. The coefficients that provided us with best model performance are $\lambda_A=1, \lambda_{SP}=0.1, \lambda_{DM}=0.5 $. For causal structure learning, we compare KGRCL with PC~\cite{spirtes2000causation}, GES~\cite{chickering2002optimal}, GIES~\cite{hauser2012characterization}, VISL ~\cite{visl2022}, NOTEARS ~\cite{notears}, DAG-GNN~\cite{DAGGNN} and a correlation graph. We evaluate the model performance by comparing the predicted graph with the ground truth graph (generated and validated by soil domain experts). We also perform an ablation study to study the effect of removing components from the KGRCL model. Here, we look at three variants, removing masking to learn only from the observed data samples that contain complete information, without the distribution matching term, and adding regression loss for estimating soil carbon.

We also present a use-case for the learned causal graphs in aiding supervised learning. This allows us to investigate the effect of integrating causal structure in prediction models. Using the causal graph as the skeleton for graph neural networks (GNN), we can add inductive biases that further inform the message-passing in the GNNs. We compare our causal graph neural networks with Random Forest, random guess, Gradient Boosted trees (GB Trees) and MLP approaches. Causal discovery methods are coupled with two variants of graph neural networks, edge-Conditioned Convolution Message Passing Neural Networks (ECMPNN) \cite{simonovsky2017dynamic, gilmer2017neural} and GraphSAGE \cite{hamilton2017inductive}. Comparing different message passing procedures allows us to study how added complexity in learning influences generalization in the prediction task. For ECMPNN, at each layer $l$ of the feed-forward neural network, the embedding signal can be computed as, 
\begin{equation}
    \mathbf{h}^l_i \leftarrow \frac{1}{|N(i)|} \sum_{j \in N(i)} F^l (E_{j, i}; W^l)h^{l-1}_j + b^l,
\end{equation}
\noindent where, $W^l$, $b^l$,$F_l$ are the weight matrix, the bias term and the neural network aggregator defined at layer $l$. On the other hand, in GraphSAGE, embeddings at search depth $k$ for given node $i$ can be computed as,
\begin{equation}
    \mathbf{h}^k_i \leftarrow \sigma ( W^k [\mathbf{h}^{k-1}_i, AGG(\{ \mathbf{h}^{k-1}_{u}, \forall u\in N(i)\})] ),
\end{equation}

\noindent where, $\sigma$ is a non-linear activation function and $W^k$ is the weight matrix at depth $k$. $\mathbf{h}^k_{N(i)} = AGG(\{ \mathbf{h}^{k-1}_{u}, \forall u\in N(i)\})$ is the signal aggregated over all sub-sampled neighbors at depth $k$. $AGG$ can be any aggregator function including trainable neural network aggregator. For both the methods, hyperparameters are fine-tuned via Bayesian optimization. 

\noindent \paragraph{\textbf{Metrics}}
For evaluating the discovered graphs, 
we use recall as an evaluation metric to compare the methods. Recall measures the proportion of ground truth edges that were accurately predicted to have existed in the generated causal graph. We also evaluate and present the precision and AUC metrics. These metrics measure if the models are accurately predicting the existence of causal edges between nodes. Moreover, as proposed in \cite{Tong2001ActiveLF}, we use the L1 edge error as another metric to compare a method's estimate with the true graph $G$. It is calculated as follows:
\begin{equation}
\begin{aligned}
      Error(P) = \Sigma_{i, j > i} I_G(X_i \rightarrow X_j)(1 - P(X_i \rightarrow X_j)) + \\ 
      I_G(X_j \rightarrow X_i)(1 - P(X_j \rightarrow X_i)) + \\
      I_G(X_i \perp X_j) P(X_i \perp X_j))
      \end{aligned}
\end{equation}
where $I_G(E)$ is an indicator function which is one, if an edge $E$ is present in the ground truth and is zero otherwise. Lower L1 edge errors suggest better causal graphs. For probabilistic approaches such as VISL, DAG-GNN, and KGRCL, the availability of probability of existence of edges allows us to compute this metric. For statistical approaches, PC, GES, GIES, we use bootstrap resampling to get an estimate of probability of existence for an edge.

Furthermore, to compare the prediction accuracy of causal GNNs with other machine learning methods, we evaluate the mean squared error and mean absolute error values. For each set of experiments, we look at multiple datasets to evaluate the generalization of the method to new locations. Two training settings are considered: the zero-shot learning scenario, where the training and test sets are from different locations (e.g., train on North Wyke and test on Washington dataset); a few-shot learning scenario, where there is some overlap between training and test sets (e.g., train on North Wyke as well as Washington and test on a holdout set from Washington). To avoid overfitting in the few-shot scenario, we ensure that the holdout set is from a different set of farms following different management practices.

%% file: results.tex

\subsection{Experimental Results}

\subsubsection{Causal Discovery}
\label{sec:res_causal_dis}

Table~\ref{tab:cd} presents the causal discovery results. We compare KGRCL with a combination of statistical and deep learning-based approaches, including  PC~\cite{spirtes2000causation}, GES~\cite{chickering2002optimal}, GIES~\cite{hauser2012characterization}, NOTEARS~\cite{notears}, DAG-GNN~\cite{DAGGNN} and VISL~\cite{visl2022}. In addition, we compare the performance of KGRCL to correlation graphs, computed by pairwise correlation between features. For these experiments, we learn causal graphs on the NW, WA and MN data sets where the results reflect causal graphs built on different number of features available for different data sets. Since, for the ground truth nodes existing the data sets, we can compare the learned graph edges with the ground truth edges. While the recall evaluates the proportion of ground truth edges predicted at a threshold of 0.5, AUC measures the overall performance aggregated over different thresholds. Several combinatorial based methods fail to predict any edges accurately leading to 0 recall and 0.5 AUC values. In contrast, KGRCL produces accurate predictions from the limited data sets available for the Washington state fields (SHOP and Home Place). For the North Wyke farmlets which include a higher number of measured variables, the performance improvement of KGRCL over other methods is smaller. Therefore, for data sets with limited sample size, KGRCL can potentially provide more improvement compared to other causal discovery methods. Comparable AUC scores from VISL in the Great Fields and Home Place fields suggest a higher generalization power than can be achieved by using continuous optimization based approaches. 

\begin{figure*}[ht]
\centering
\begin{tabular}{cccccc}
\hspace{-1pt}
 \subcaptionbox{\label{fig:adj-gt} Ground Truth  \vspace{5pt} }{\includegraphics[scale=0.2]{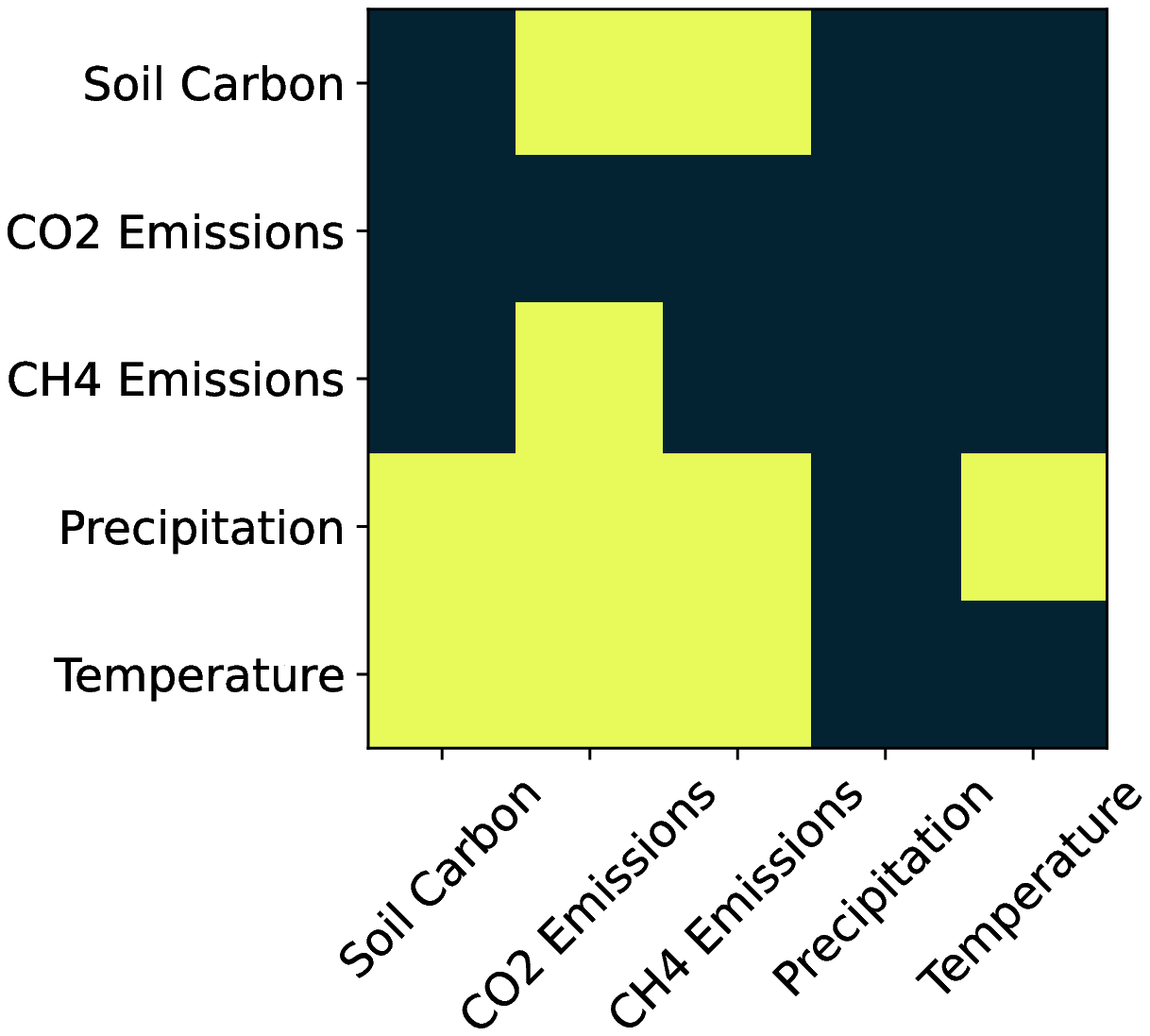}} &
 \subcaptionbox{\label{fig:adj-pc} PC Algorithm  \vspace{5pt} } {\includegraphics[scale=0.2]{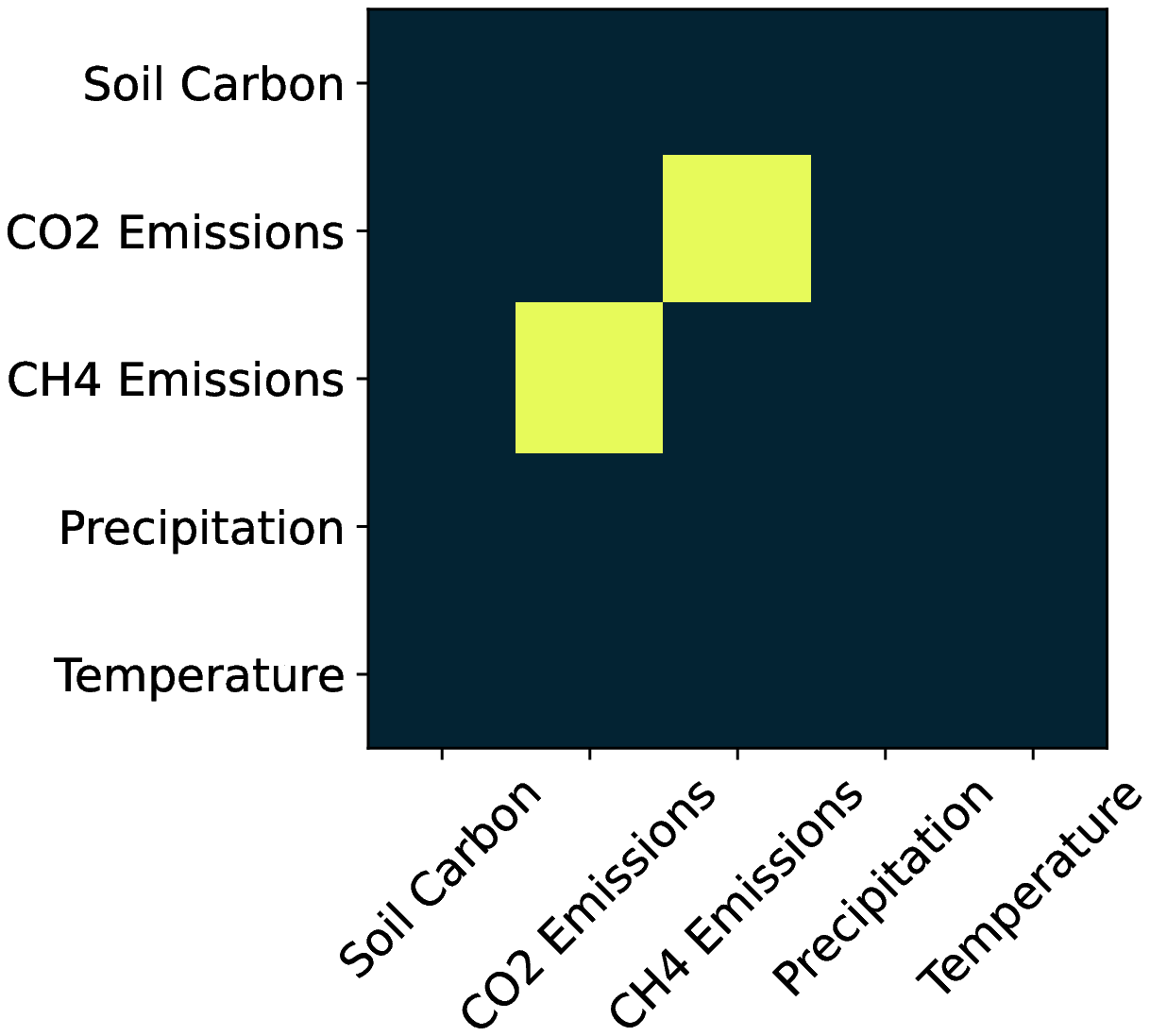}}  &
 \subcaptionbox{\label{fig:adj-ges} GES Algorithm \vspace{5pt} }{\includegraphics[scale=0.2]{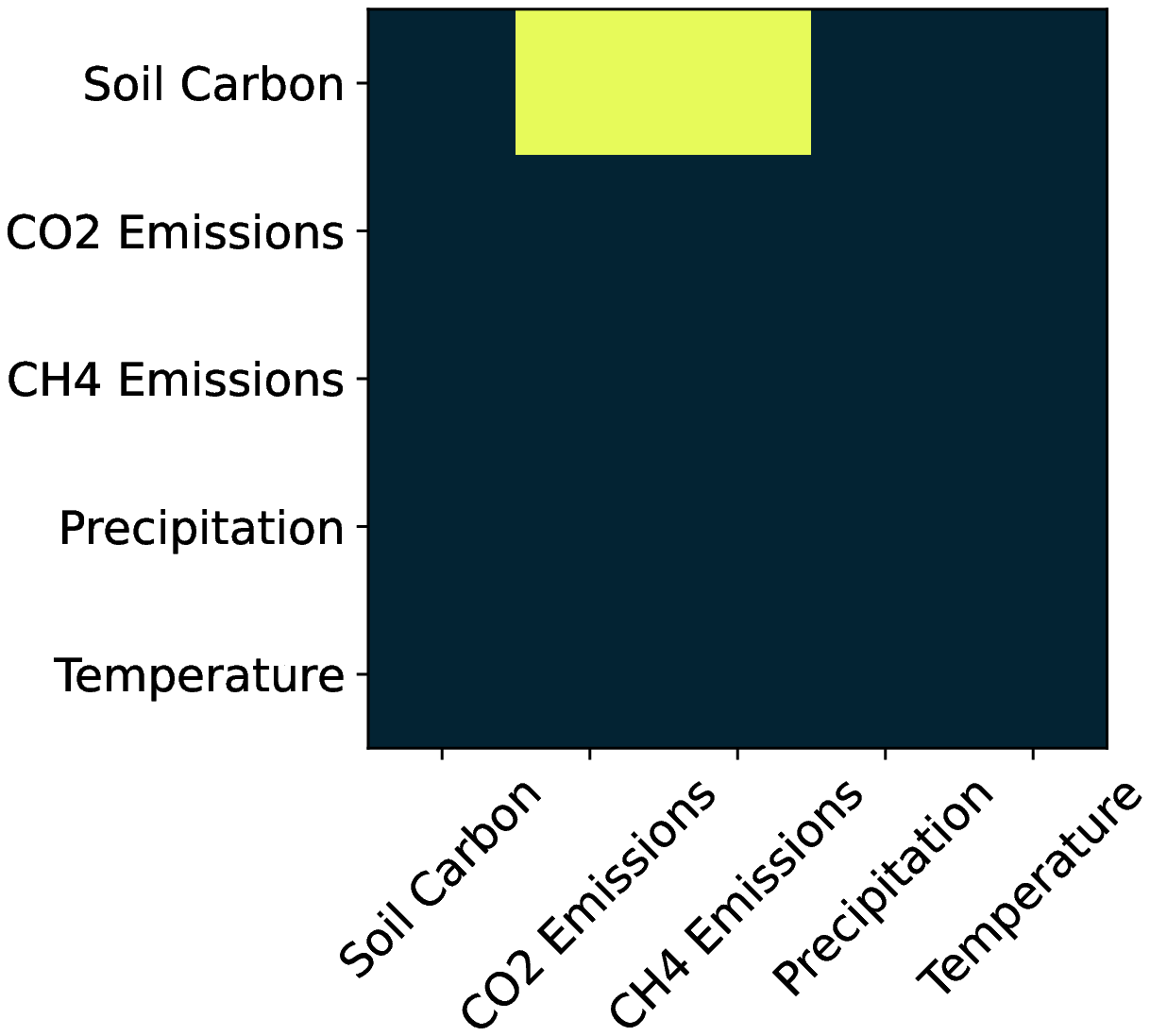}} &
 \subcaptionbox{\label{fig:adj-gies} GIES Algorithm \vspace{5pt} }{\includegraphics[scale=0.2]{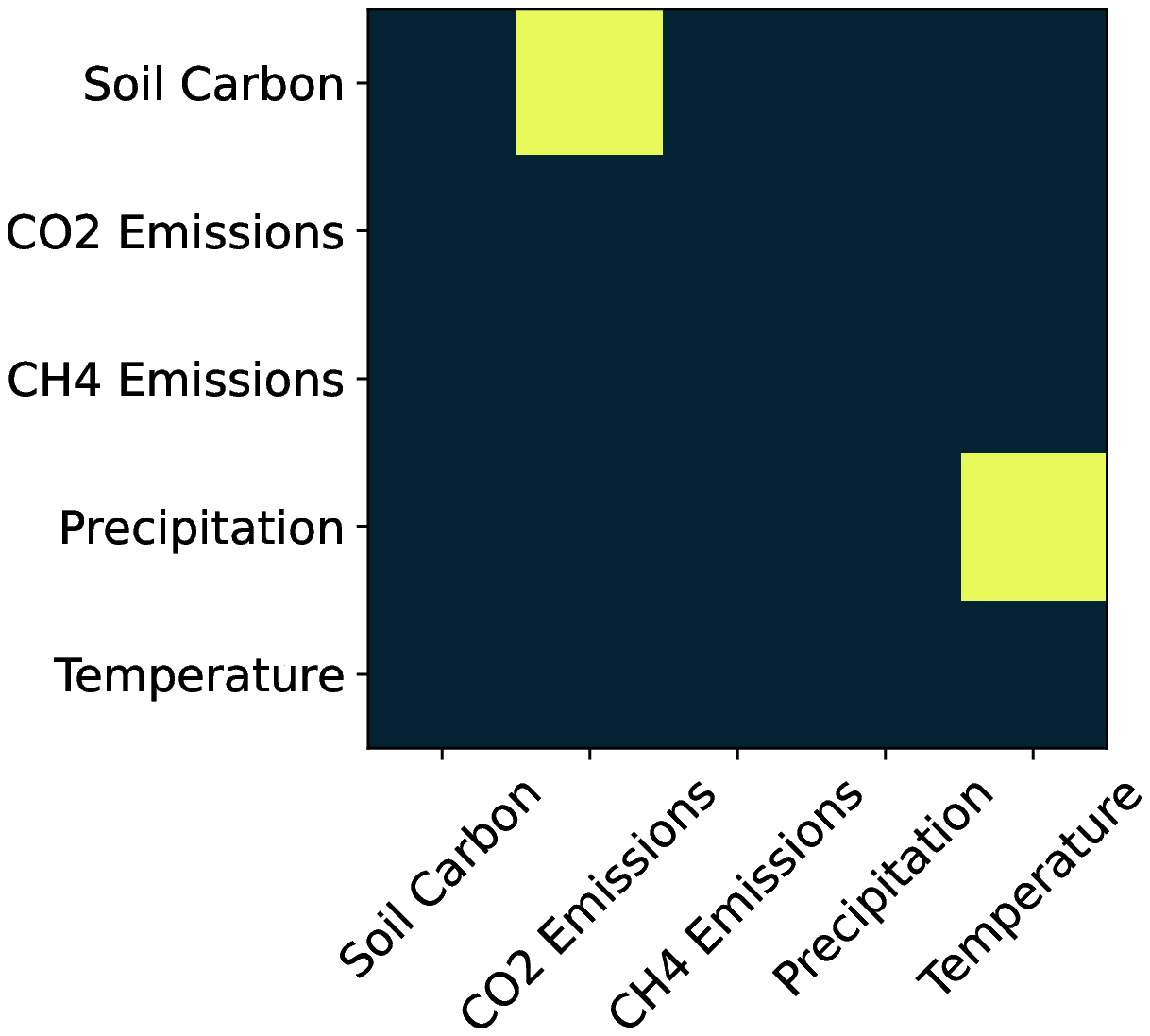}} &
 \subcaptionbox{\label{fig:adj-visl} VISL \vspace{5pt} }{\includegraphics[scale=0.2]{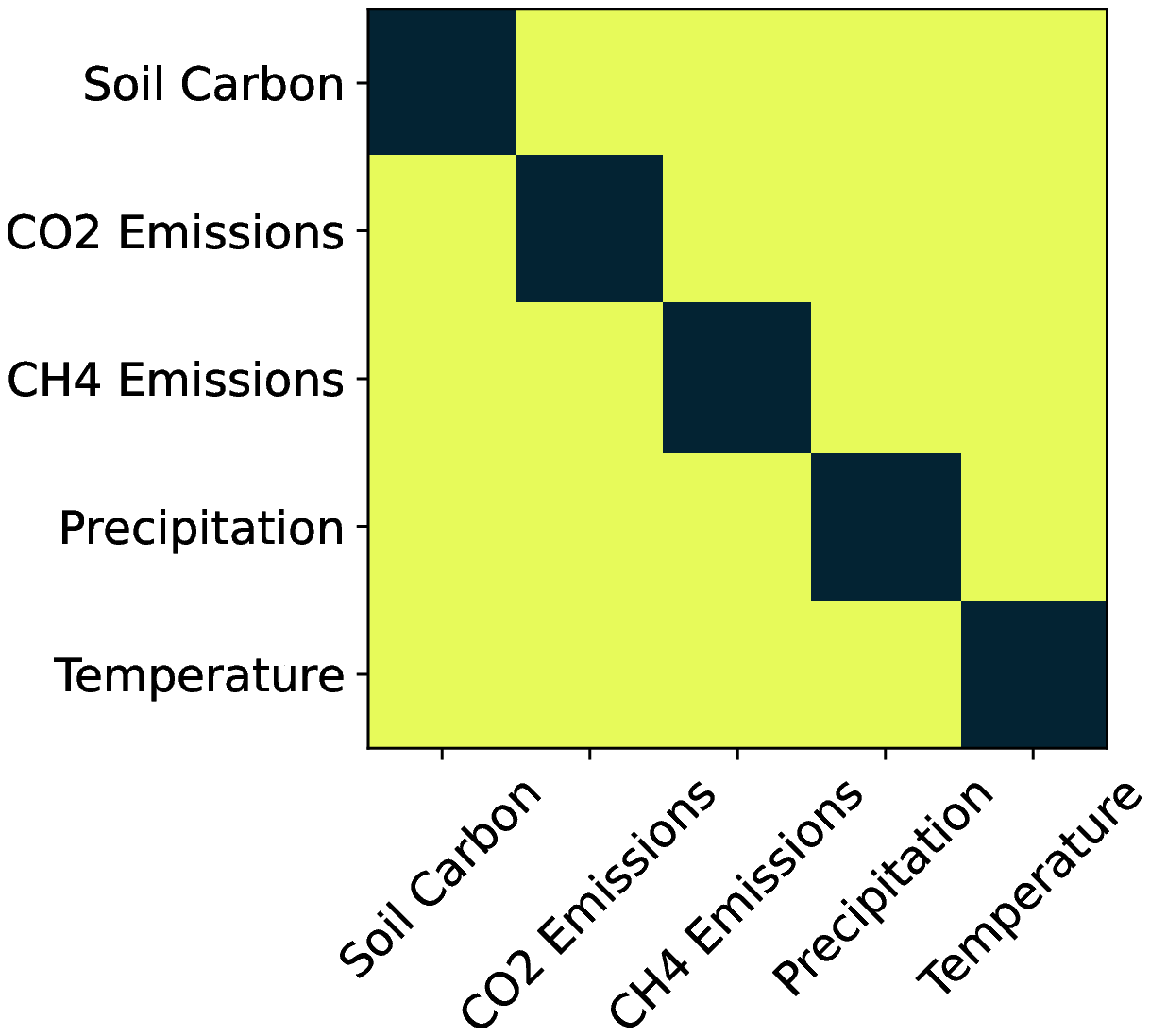}} &
 \subcaptionbox{\label{fig:adj-kgrcl} KGRCL \vspace{5pt} }{\includegraphics[scale=0.2]{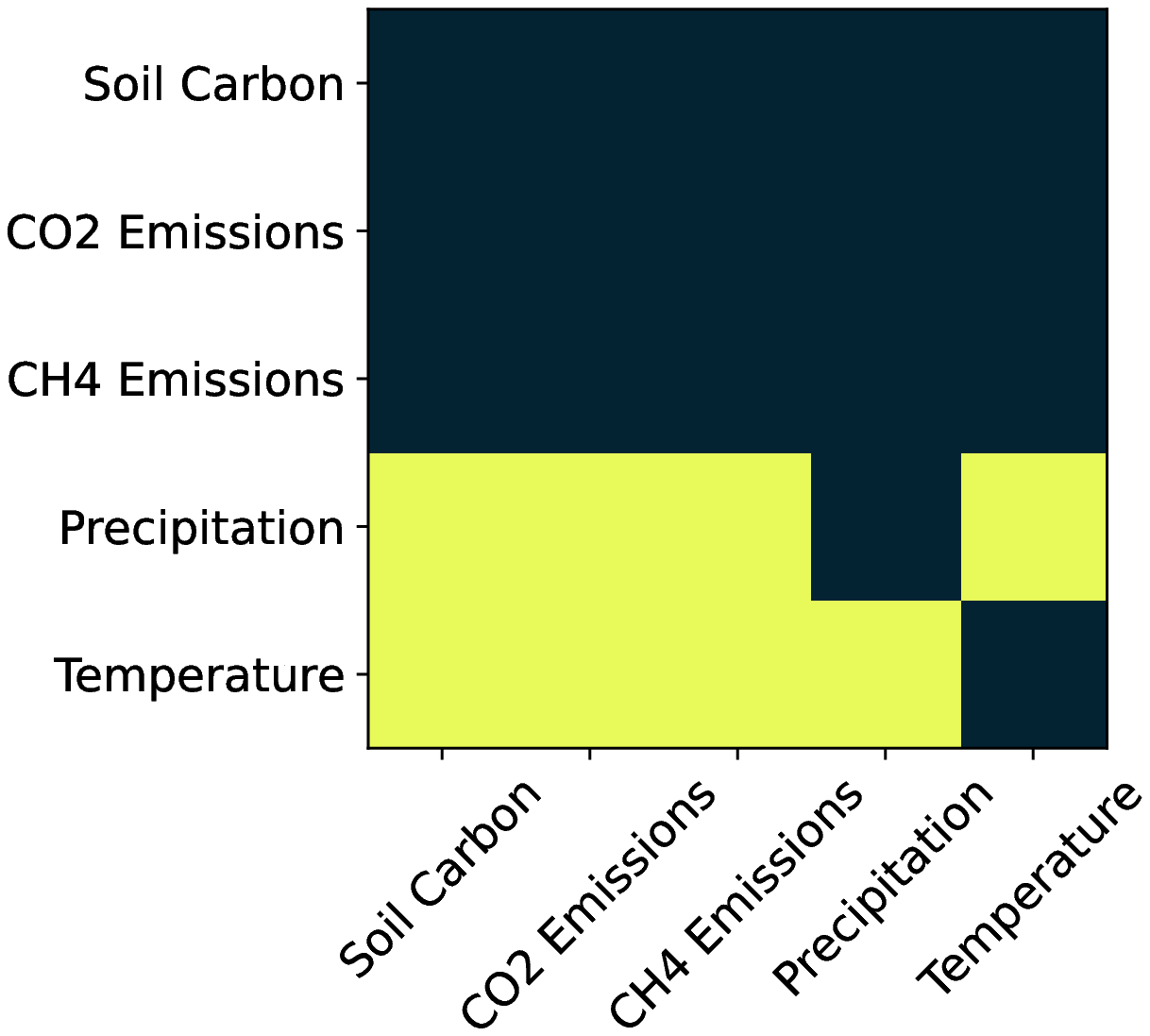}} \\ 
\end{tabular}
\vspace{-10pt}
\caption{ Adjacency Matrices for nodes included in ground truth graph for NW data sets. Adjacency matrix represents edge existence for row nodes causing column nodes. Yellow represents existence of edge. For KGRCL and VISL, edge existence probabilities are thresholded at 0.5 to obtain edge existence. }
\label{fig:adjacency_matrix}
\vspace{-10pt}
\end{figure*}

\begin{figure*}
    \centering
    \includegraphics[scale=0.4]{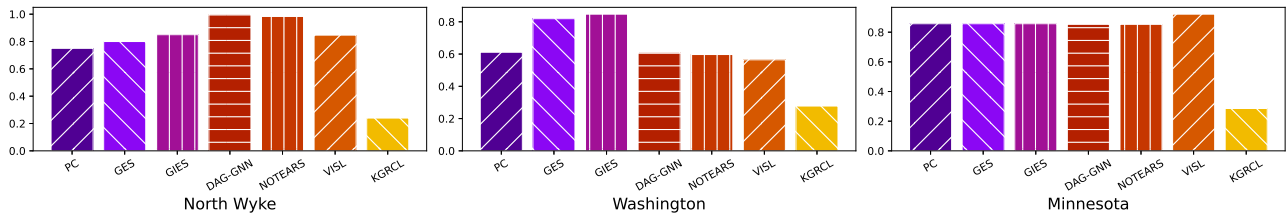}
    \caption{Standardized L1 Edge Error. For each data set we divide the L1 edge error by the maximum number of edges that can exist in the ground truth graph. }
    \label{fig:l1_edge_error}
\end{figure*}

Furthermore, we can visually compare graphs learned by the different methods. Figure~\ref{fig:adjacency_matrix} compares the causal graphs learned on the North Wyke data set. We can see that PC, GES and GIES predict a sparse graph while VISL estimates the existence of all edges. We can also compare the causal discovery methods in terms of the L1 edge error. Figure~\ref{fig:l1_edge_error} shows that KGRCL is able to estimate both the existence and the non-existence of edges more accurately than other methods in terms of the L1 edge error scores.

\begin{figure*}
       
      \subfloat[w/o DM for NW]{\label{fig:before-acdm-nw}    \includegraphics[width=0.45\linewidth]{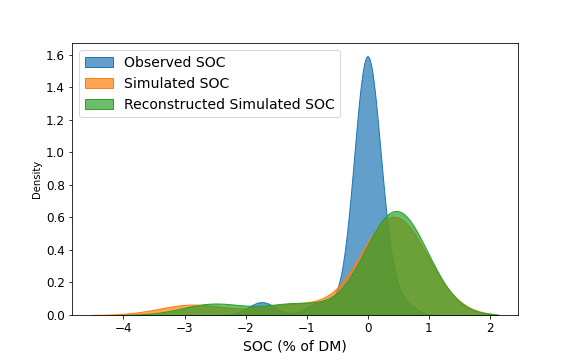} }
    \subfloat[w DM for NW]{\label{fig:after-acdm-nw}
    \includegraphics[width=0.45\linewidth]{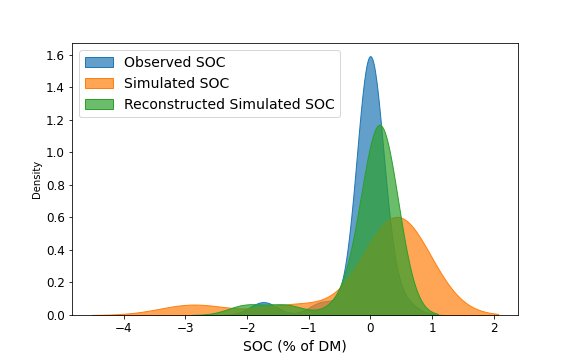}
    }
  \caption{Effect of conditional distribution matching on simulated data. Soil carbon simulated data representations without (w/o) and with (w) distribution matching (DM) for North Wyke fields (NW). }
  \label{fig:rl-plots}
        \end{figure*}

Apart from the structure learning results, we can also evaluate the distributional shift in simulated data. For instance, Figure~\ref{fig:rl-plots} (and Appendix, Figure~\ref{fig:rl-plots-all}) compares simulated soil carbon values reconstructed from KGRCL without adding the distribution matching loss term (w/o DM) and with the distribution matching loss term (w DM) for the NW data set. 
We can notice that the overestimation of simulated soil carbon values can be rectified by learning better representations using observed data. The density plots reflect a shift in the distribution of the reconstructions that better emulate the observed values.

\subsubsection{Ablation Study}
\label{sec:res_ablation}
\begin{figure*}
\scriptsize
    \centering
    \subfloat[AUC]{\label{fig:auc}    \includegraphics[scale=0.3] {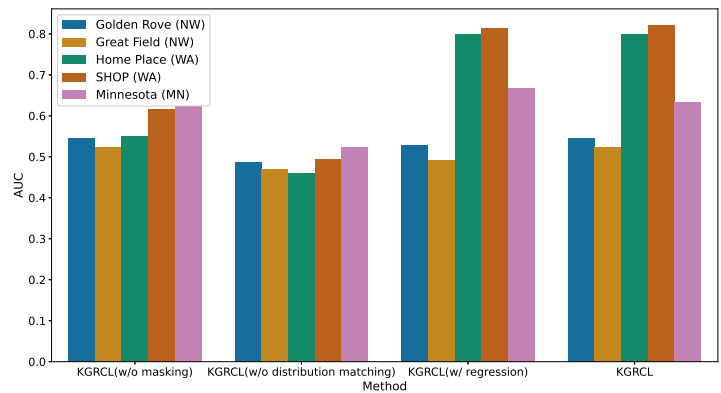} }  
    \hspace{-0.1pt}
    \subfloat[Recall]{\label{fig:recall}
    \includegraphics[scale=0.3]{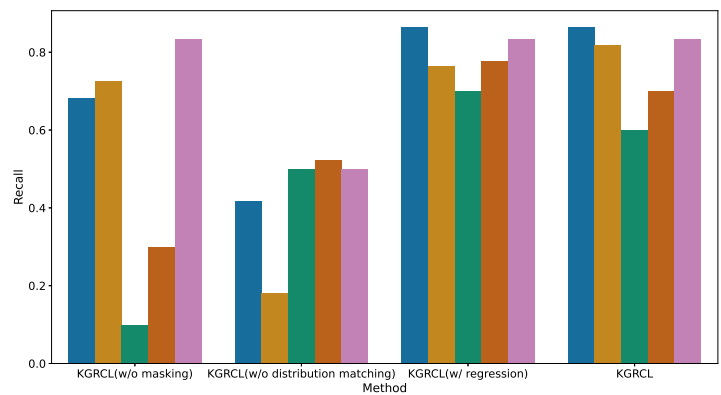}
    }
     \caption{Ablation study of aspects of KGRCL such as the effect of distribution matching on AUC and Recall of the discovered graph edges. }
         \label{fig:ablation}
   \end{figure*}

We also include an ablation study comparing different KGRCL framework variants in Figure~\ref{fig:ablation}. Since there are several missing values in the observed data used for model learning, we can mask the loss function, focusing instead upon available observed samples. In the ablation study, we investigate the effect of removing the mask. In this case, a feed-forward network is used to impute missing variables using those from observed and simulated data where the information is available for all samples. Without masking the loss function, the representations and causal discovery learn the bias from the imputed values as well, reducing the model performance. We may also investigate the effect of removing the $\mathbf{\text{Loss}_{\text{DM}}}$ on causal structure learning. The conditional distribution matching loss term not only provides feedback for the simulated representations but also indirectly influences learning of other parameters in the deep learning framework. The results without distribution matching reflect a decline in model performance on the learned causal structure for the observed data set as well. We investigated the impact of adding $\mathbf{\text{Loss}_{\text{SP}}}$ further. Providing more penalty on errors arising from incorrect predictions in observed soil carbon values, the model is able to learn the behavior of observed soil carbon data over the simulated ones. This provides improvement in model performance for the Washington and Minnesota sites.

\subsubsection{Supervised learning for Soil carbon prediction}
\label{sec:res_causal_gnn}
Table~\ref{tab:causalgnn} and Table~\ref{tab:causalgnn-fewshots} compares ML methods for the soil carbon prediction task. To evaluate model performance in out-of-distribution scenarios, we split the train and test data by different spatial locations with different land use types. For a systematic comparison of model performance across the data sets, the ML models learn to estimate soil carbon using features that are common among all data sets. This includes air temperature, precipitation, soil silt fraction, soil sand fraction, bulk density, pH, fertilizer use and soil carbon values.

Table~\ref{tab:causalgnn} presents the zero-shot learning results. The causal graph based GNN models are able to provide better prediction accuracy than conventional ML models. The choice of train and test site also greatly affects the model performance. For instance, compared to other data sets, Minnesota (MN) data sets was collected in a more controlled environment. Therefore, the models learned on Minnesota data and tested on other sites result in higher prediction errors due to the distributional shift between the train and test data sets. We also see a difference in model performance between the GraphSAGE and ECMPNN architectures in the case of distributional shift between the train and test data sets. Since GraphSAGE samples node neighborhoods, the model is unable to leverage the global graph structure for effective learning. Therefore, in the presence of larger distributional shifts, there is a contrast in the model performance of GraphSAGE and ECMPNN. Among the GNN based methods, KGRCL based skeletons improve the prediction error. While learning from a local graph structure enables KGRCL+GraphSAGE model to outperform when trained and tested using NW and WA data sets, KGRCL+ECMPNN achieves a lower prediction error. Similarly, the few-shots learning results presented in Table~\ref{tab:causalgnn-fewshots}, we see a difference in model performance for GraphSAGE and ECMPNN when there is a distributional shift between the test and train data sets. In experiments where we test on Minnesota holdout set, there is a noticable difference between the train and test data sets. Similarly, in the experiments where we test on the NW holdout set, the differences in management practices between the NW train and holdout set are reflected in the higher evaluation metrics for GraphSAGE model. We see that improved prediction accuracy can be achieved by using causal GNN based approaches. In both the zero-shot learning setting and the few-shot learning setting, KGRCL based GNNs provide the best model performance in terms of the evaluations metrics.


\begin{table*}
   \centering
   \scriptsize
    \resizebox{\textwidth}{!}{
    \begin{tabular}{p{2cm}llllllllll}
        \hline
        \multirow{2}{*}{\bfseries  } & 
        \multicolumn{2}{c}{\bfseries Golden Rove, NW} & 
        \multicolumn{2}{c}{\bfseries Great Field, NW} & 
        \multicolumn{2}{c}{\bfseries HomePlace, WA} & 
        \multicolumn{2}{c}{\bfseries SHOP, WA} & 
        \multicolumn{2}{c}{\bfseries MN} 
        \\ 
        \cmidrule(lr){2-3}
        \cmidrule(lr){4-5}
        \cmidrule(lr){6-7}
        \cmidrule(lr){8-9}
        \cmidrule(lr){10-11}
        Method & Recall & AUC & Recall & AUC & Recall & AUC& Recall & AUC& Recall & AUC \\ 
        \cmidrule(lr){1-11}
        KGRCL     & \textbf{0.8636}   & \textbf{0.5444} & \textbf{0.8181} & 0.5246 & \textbf{0.6000} & \textbf{0.8000} & \textbf{0.7000} & \textbf{0.8200} & \textbf{0.8333} & \textbf{0.6333} \\
        PC     &  0.0454    & 0.5142  & 0.0000  & 0.5000 & 0.1000  & 0.5166& 0.0000  & 0.5000& 0.0000& 0.5000 \\
        GES  &   0.0454    & 0.5227  & 0.0454    & 0.5227&0.000  & 0.5000& 0.0000  & 0.5000& 0.0000& 0.5000 \\
        GIES & 0.0000      & 0.4915 & 0.0454  & 0.5227 & 0.1000  & 0.5000& 0.0000  & 0.5000& 0.0000& 0.5000 \\
        NOTEARS & 0.0909 & 0.5177 & 0.0454    & 0.4984 & 0.3333  & 0.2222 & 0.3333  & 0.3333& 0.3333 & 0.5000 \\
        DAG-GNN & 0.0000  & 0.4745 & 0.0000  & 0.5000 & 0.0000  & 0.5000 & 0.0000  & 0.5000& 0.0000 & 0.5000 \\
        VISL & 0.3181 & 0.5023 & 0.2727  & \textbf{0.5285} & 0.5000  & 0.5533 & 0.5000   & 0.4866& 0.6667 & 0.5244 \\
        Correlation& 0.2272  & 0.4791 & 0.0000  & 0.4661  & 0.3000  & 0.3666 & 0.5000  & 0.3333& 0.6667 & 0.1667 \\   
        Graph&&&&&&&&&& \\
        \hline
    \end{tabular}
    }
    \caption{Causal discovery model performance. KGRCL is able to outperform other popular causal discovery methods.}
    \label{tab:cd}
\end{table*}

\begin{table*}
    \centering
        \small
    \begin{tabular}{lllllllll} 
        \hline
        \multirow{1}{*}{\bfseries Train Site} & 
        \multicolumn{2}{c}{\bfseries WA} & 
        \multicolumn{2}{c}{\bfseries NW} & 
        \multicolumn{2}{c}{\bfseries NW, WA} &
        \multicolumn{2}{c}{\bfseries MN}\\ 
        \multirow{1}{*}{\bfseries Test Site } & 
        \multicolumn{2}{c}{\bfseries NW} & 
        \multicolumn{2}{c}{\bfseries WA} &
        \multicolumn{2}{c}{\bfseries MN} & 
        \multicolumn{2}{c}{\bfseries NW, WA}\\ 
        \hline
        Method & MSE & MAE & MSE & MAE & MSE & MAE & MSE & MAE \\ 
        \hline
        Random Guess  & 1.2321&0.8821&1.1604&0.8749& 1.2703 & 0.9041& 1.4953 & 0.9752 \\
        \hline
        Random Forest & 0.1287 &0.3190&0.1296&0.3181& 0.2631 & 0.5129 & 0.2631 & 0.5129 \\
        GB Trees  & 0.1239& 0.3120 &0.1135&0.3009& 0.2542 & 0.5142 & 0.2452 & 0.5042\\
        MLP  & 0.1231&0.3142&0.1036&0.1969& 0.3062 & 0.4906 & 0.2210 & 0.4694 \\    
        \hline
        PC + GraphSAGE    & 0.0989&0.2441&0.1104&0.2888 & 0.0920 & 0.3033 & 0.2011 & 0.4432\\
        PC + ECMPNN    & 0.0790&0.2253&0.1114&0.3391& 0.0108 & 0.4093& 0.0316 & 0.4570 \\
        GES + GraphSAGE  & 0.0881&0.2104&0.1137&0.3145& 0.0692 & 0.2630 & 0.2057 & 0.4541\\
        GES + ECMPNN  & 0.0680&0.2276&0.1192&0.3093& 0.0217 & 0.3472 & 0.0277 & 0.4318\\        
        GIES + GraphSAGE & 0.0855&0.2143&0.0878&0.2449& 0.0978 & 0.2265& 0.2062 & 0.4531 \\        
        GIES + ECMPNN & 0.0718&0.2204&0.1162&0.3176& 0.0472 & 0.4990 &0.0202 & 0.2756\\
        \hline
        \textbf{KGRCL + GraphSAGE}     &\textbf{0.0566}&\textbf{0.1012}&\textbf{0.0603}&\textbf{0.1949} & 0.0575 & 0.2399 & 0.1160  & 0.2721\\
        \textbf{KGRCL + ECMPNN}     & 0.0816&0.2350&0.0862&0.2318 & \textbf{0.0027} & \textbf{0.2112}& \textbf{0.0069} & \textbf{0.2456} \\
        \hline
    \end{tabular}
    \caption{Performance comparison for soil carbon prediction in out-of-sample scenarios. The three sites North Wyke, Washington and Minnesota differ in soil textures, management practices and weather patterns. }
    \label{tab:causalgnn}
\end{table*}

\begin{table*}
    \centering
    \small
    \begin{tabular}{lllllllllll} 
        \hline
        \multirow{1}{*}{\bfseries Train Site  } & 
        \multicolumn{2}{c}{\bfseries NW, WA } & 
        \multicolumn{2}{c}{\bfseries NW, MN } & 
        \multicolumn{2}{c}{\bfseries NW, MN , WA}& 
        \multicolumn{2}{c}{\bfseries NW, MN, WA}& 
        \multicolumn{2}{c}{\bfseries NW, MN, WA} \\ 
        \multirow{1}{*}{\bfseries Test Site } & 
        \multicolumn{2}{c}{\bfseries WA } & 
        \multicolumn{2}{c}{\bfseries MN } &
        \multicolumn{2}{c}{\bfseries WA }&
        \multicolumn{2}{c}{\bfseries MN }&
        \multicolumn{2}{c}{\bfseries NW }\\ 
        \hline
        Method & MSE & MAE & MSE & MAE & MSE & MAE  & MSE & MAE& MSE & MAE \\ 
        \hline
                Random Guess  &1.0803 & 0.8277 & 1.0643 & 0.8309& 1.1593&0.8607&1.2289&0.8816& 1.4086& 0.9254 \\
                \hline
        Random Forest &0.0516 & 0.1897 & 0.0110 & 0.1656 &0.0684&0.1823&0.1851&0.4302& 0.1204 & 0.3402 \\
        GB Trees  &0.0545 & 0.1897 & 0.0570 & 0.1753& 0.0754&0.1912&0.1427&0.3778& 0.1345 & 0.3583\\
        MLP  &0.0578 & 0.1897 & 0.0180 & 0.1656 &0.0742&0.1948&0.1042&0.4086& 0.0762 & 0.3801 \\  
        \hline
         PC + GraphSAGE    &0.0654&0.2239&0.0336&0.0301 & 0.0813 & 0.2247 & 0.2057 & 0.4536& 0.1833 & 0.4223 \\
                PC + ECMPNN    &0.0582 & 0.1900 & 0.0249 & 0.0225 & 0.0609&0.1863&0.0786&0.2703& 0.1192 & 0.4139 \\

        GES + GraphSAGE  &0.0554&0.2239&0.0335&0.0116 & 0.0963 & 0.2676 & 0.2057 & 
        0.4536& 0.1833 & 0.4215 \\
        GES + ECMPNN  &0.0579&0.2126&0.0221&0.0986 & 0.0859 & 0.1866 & 0.0902 & 0.1164& 0.0760 & 0.3713\\
        GIES + GraphSAGE &0.0739&0.2448&0.0057&0.0753 & 0.0886 & 0.2542 & 0.2053 & 0.4531& 0.1137 & 0.3472 \\

        GIES + ECMPNN &0.0407&0.1600&0.0241&0.0539 & 0.0665 & 0.1596 & 0.0119 & 0.1182& 0.1341 & 0.3668 \\
        \hline
        \textbf{KGRCL + GraphSAGE }    &\textbf{0.0001}&\textbf{0.0062}&0.0015& \textbf{0.0108} & \textbf{0.0001} & \textbf{0.0041} & 0.0107 & 0.0977 & 0.0629 & 0.1902 \\
       
        \textbf{KGRCL + ECMPNN  }   &0.0012&0.0093&\textbf{0.0003}&0.0109 & 0.0010 & 0.0420 & \textbf{0.0015} & \textbf{0.0146}& \textbf{0.0570} & \textbf{0.1876}\\

        \hline
    \end{tabular}
    \caption{Performance comparison for soil carbon prediction in few-shot learning. The three sites North Wyke, Washington and Minnesota differ in soil textures, management practices and weather patterns. }
    \label{tab:causalgnn-fewshots}
\end{table*}

    

%% file: discussion.tex
\section{Discussion}
\label{sec:discussion}
We use a deep learning based causal structure learning method employing both simulated and observed data. In the framework, we incorporate conditional distribution matching to refine the simulated data distribution as an effective merging and learning strategy from the two data sources. The causal structure learning approach is easily scalable to a few hundred features. Our empirical results show that distribution matching improves the simulated process-model data to match observed data. It also indirectly influences the parameter estimates for the observed data. In essence, the results of our ablation studies show that we achieve a superior causal graph by employing distribution matching. 

We discussed a case study on soil organic carbon, which is one of the principal factors affecting agricultural productivity and also has a co-benefit of managing climate change. Accurately estimating and managing soil organic carbon is thus imperative to solve long term sustainability goals. 
Currently, the soil community widely uses  process based models for soil organic carbon content prediction. 
One of the challenges is the applicability of process models to out-of-distribution settings such as changing weather, soil textures and management practices. In this context, the transportability of causal approaches to unseen scenarios is critical because of low data availability. Soil data is typically collected at a low frequency and few parameters affecting soil organic carbon are observed. By incorporating multiple data sources such as process-based model data, we can access missing confounders that are not present in the observed data. This reduces the need for extensive data collection. It also solves the need for large number of data samples required by deep learning approaches. 
Using our framework, we are able to discover relationships among process that are either well-established in the soil science literature and can help by bringing forth the connections between the physical, chemical and structural properties of soil. It also allows better interpretability of soil at the system's level. To highlight the complexity of the problem, a partial graph including some processes and management practices is shown in Appendix ~\ref{fig:kgrcl_nw_graph}. We notice that there is a strong association between Total Carbon (\% of DM) and Total Nitrogen (\% of DM) in soil and soil organic matter (SOM). These relationships are consistent with theory and other studies~\cite{CtoN}. Also, note that fertilizer addition adds more total Nitrogen~\cite{MENG20052037}. Other well known relationships we discover are that ploughed soils change the pH of the soil, it increases the bulk density, and it reduces soil organic carbon~\cite{tillage_ph}. In addition, we discover relationships across simulated and observed data, such as fertilizer addition (observed) affects NO3-leaching (nitrate loss due to leaching), NO-flux (daily Nitric Oxide emission rate) and, Litter\_N (daily litter incorporation), which are process model outputs. Other uses of causal structure learning is using the learned graph in structural causal models for causal effect estimation. This can help guide farmers to use sustainable farming practices that have a positive effect on soil health. Causal graph analysis opens new avenues for soil scientists and domain experts to approach the problem area from new dimensions and form novel hypothesis to test experimentally.   

Recent work ~\cite{BRUNGARD2021114998} highlighted that there are regional differences in accuracy when estimating soil organic carbon using a global model. They showed that region specific and ensemble models were less uncertain than global models. To learn more about this, we studied the performance of different prediction models on new farm locations in two out-of-sample scenarios that arise in practical situations. One is when there is a need to make predictions for a new farm with no data (zero-shot learning) and another where the test farm has few data points (few-shot learning). 
Because of the complexity of soil mechanisms, there are still unobserved confounding variables such as differences in microbial nature and activity that are region specific. Furthermore, this could modify the effect of certain soil processes depending on factors such as local weather, which affect the edge probabilities. The causal graph estimated by our algorithm, KGRCL, outperforms other causal and non-causal approaches in both zero-shot and few-shot scenarios. We show that by using local data a lower prediction error can be achieved, especially when few observed samples are available for a test site. We achieve the lowest error of 0.0041 \% of dry matter for soil carbon prediction for the WA site using the common features across data sets. These common features are part of the regular soil sampling and data collection that farms typically perform yearly. If farmers collect data (including soil pH, soil organic carbon, bulk density, soil texture, cation exchange capacity, farm management data and weather parameters), causal approaches can help them understand the state of their soil's organic carbon and eventually the use of causal effect inference can allow them to choose more sustainable practices to conserve organic carbon. 

Currently, we provide validation for a limited number of graph edges. Some of these are novel discoveries that are not well studied in the literature so they are easily verifiable. In the future, we plan on validating the larger estimated causal graph validation through a combination of deductive reasoning using large language models and using expert knowledge. The supervised prediction was performed only on the common features across the data sets. Future work will entail 
exploring causal fusion approaches~\cite{causal_fusion} to effectively query multiple data sets where each dataset is collected under heterogenous conditions (such as differences in sampling, collection and analysis methods), and be flexible to future changes in data collection.

%% file: related_work.tex
\section{Related Work}
Several causal discovery methods have been proposed in the literature for learning a DAG from the data. Popular statistical approaches include the PC algorithm, a  constraint-based method,  ~\cite{spirtes2000causation} and score-based methods such as, Greedy Equivalence Search (GES) ~\cite{chickering2002optimal, meek1997graphical} and Greedy Interventional Equivalence Search (GIES) ~\cite{hauser2012characterization}. Recently, deep learning approaches have been leveraged to learn the causal graphs due to their scalability and their ability to capture non-linear relationships among covariates. Furthermore, using a neural network based learning method offers us more flexibility in terms of not being tied to specific model assumptions. In addition, the ability to perform global updates on our parameter set frees the model from restricting assumptions about the local graph structure \cite{notears, geffner2022deep, visl2022}. Variational autoencoder and graph neural network-based architecture are among the widely used methods for causal structure learning~\cite{DAGGNN, visl2022, ignavier-ae, yang2021causalvae, kocaoglu2017causalgan}. More detailed discussion on causal discovery methods can be found in ~\cite{spirtes2016causal}. 


 

%% file: conclusion.tex
\section{Conclusion}
In this paper, we present the KGCRL framework for causal structure learning and representation learning. The framework enables us to learn causal graphs using both simulated and observed data. This can be useful in improving understanding of natural and engineered systems in scientific domains. In our experiments, we calibrate simulations using regional, observed datasets that can further be used by farmers and researchers to study changes in soil processes in agricultural farms. Furthermore, we demonstrate that the learned causal relationships improve prediction accuracy in out-of-distribution soil carbon prediction tasks. Our results show that a lower prediction error can be achieved with few-shot learning where we inform the causal model with local farm data.  The framework can be generalized to other applications in scientific domains where simulated data can be refined using observed data.

%% file: appendix.tex
\section{Appendix}

\subsection{Causal discovery implementations}
For reproducibility purposes, we provide the implementations we used of the causal discovery algorithms we compare with:
\begin{itemize}
    \item We use the PC, GES and GIES implementation from the doWhy library~\cite{dowhy}.
    \item The VISL implementation is available here (~\url{https://github.com/microsoft/causica}.
    \item NOTEARS implementaton is available here (\url{https://github.com/xunzheng/notears}).
    \item We use the following DAG-GNN implementation ~\url{https://github.com/ronikobrosly/DAG_from_GNN}/
\end{itemize}

\begin{figure*} 
\centering
\begin{tabular}{ccccc}
 \subcaptionbox{\label{fig:before-acdm-nw1} w/o DM, NW  
 } 
 {\includegraphics[scale=0.16]{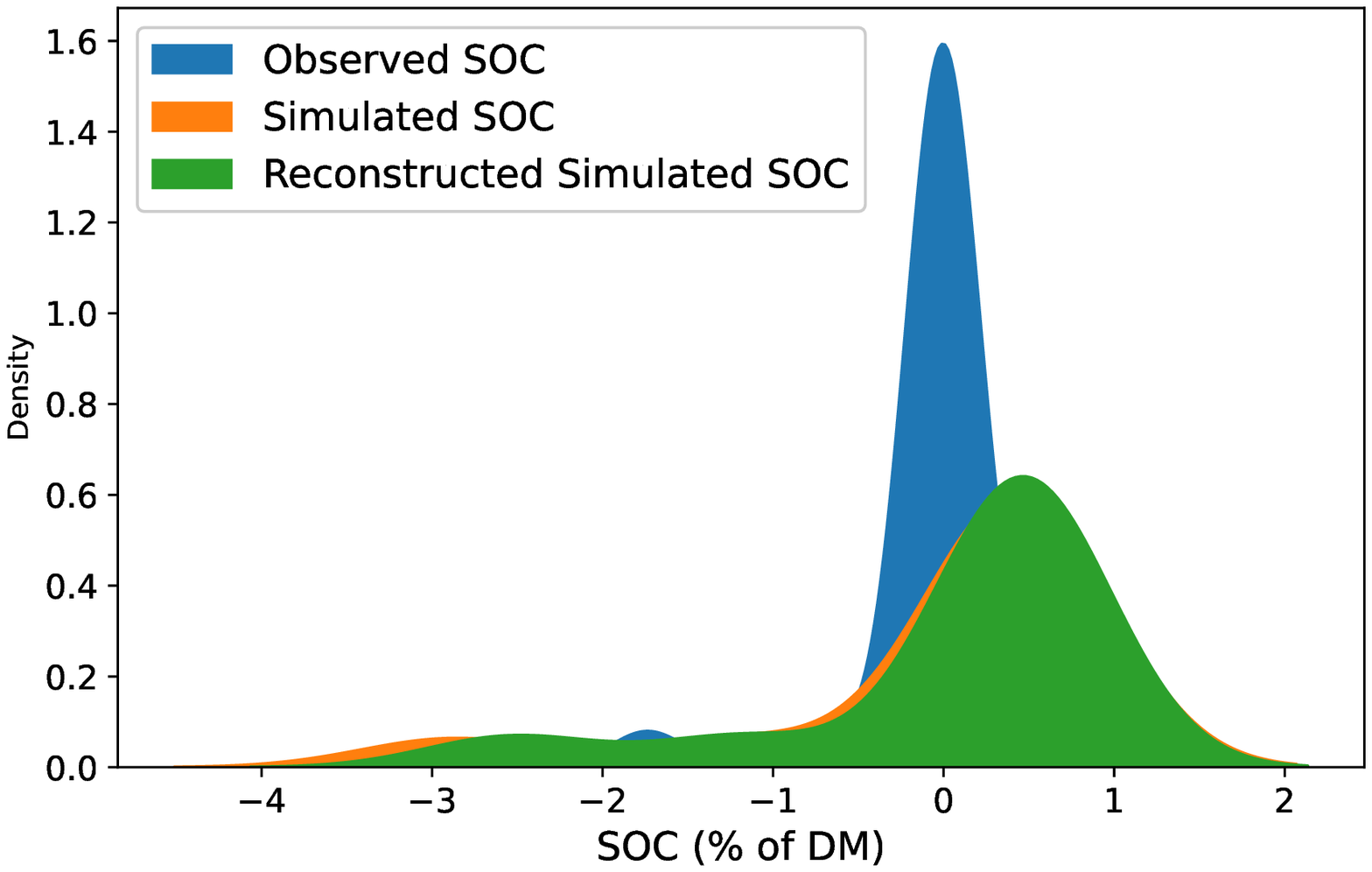}} &
  \subcaptionbox{\label{fig:before-acdm-nw-gr} w/o DM, GR, NW  \vspace{5pt} }{\includegraphics[scale=0.16]{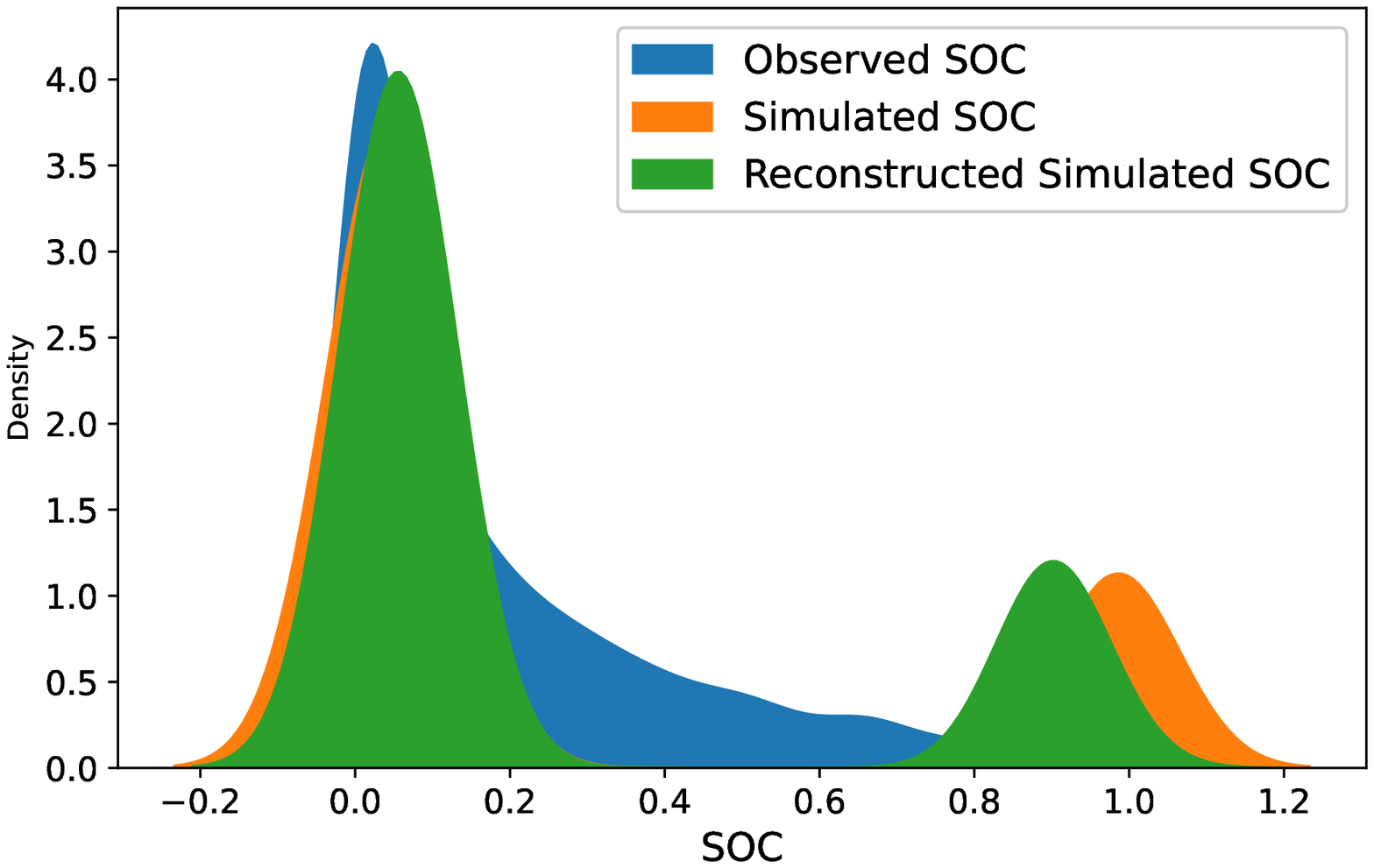}} &
  \subcaptionbox{\label{fig:before-acdm-nw-gf} w/o DM, GF, NW  \vspace{5pt} }{\includegraphics[scale=0.16]{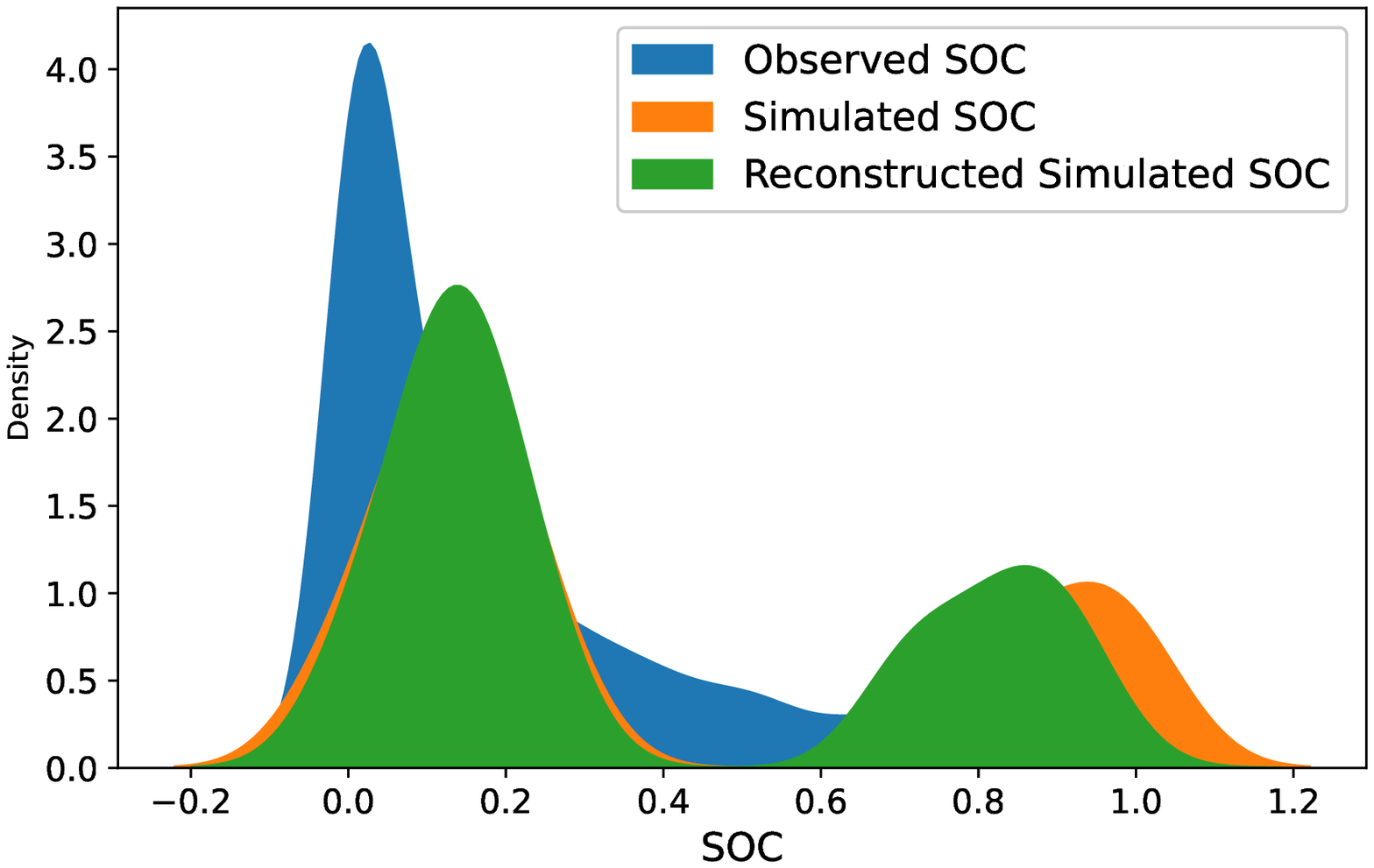}} &
  \subcaptionbox{\label{fig:before-acdm-wa-shop} w/o DM, SHOP, WA  \vspace{5pt} }{\includegraphics[scale=0.16]{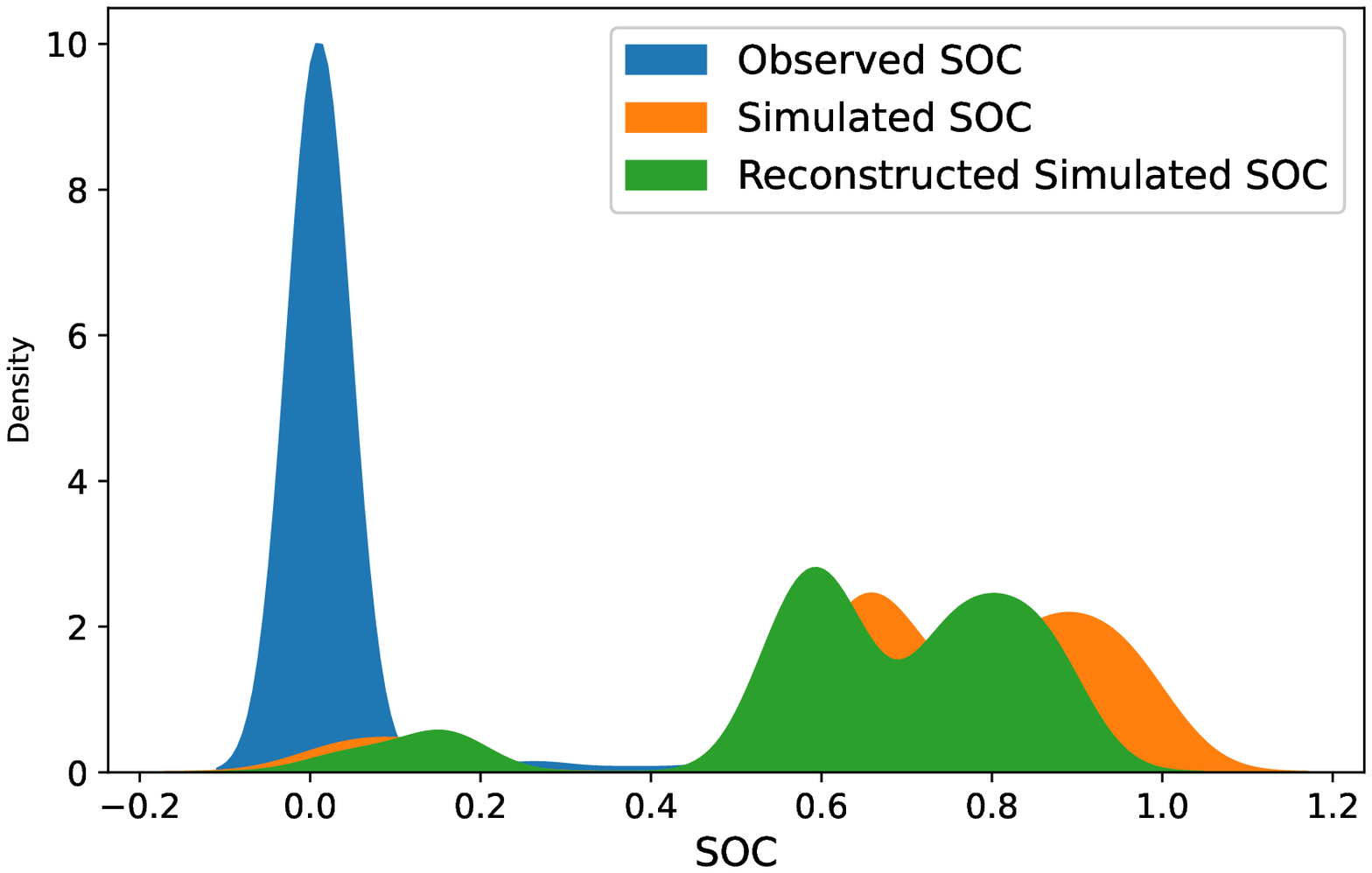}} &
  \subcaptionbox{\label{fig:before-acdm-wa-hp} w/o DM, HP, WA \vspace{5pt} }{\includegraphics[scale=0.16]{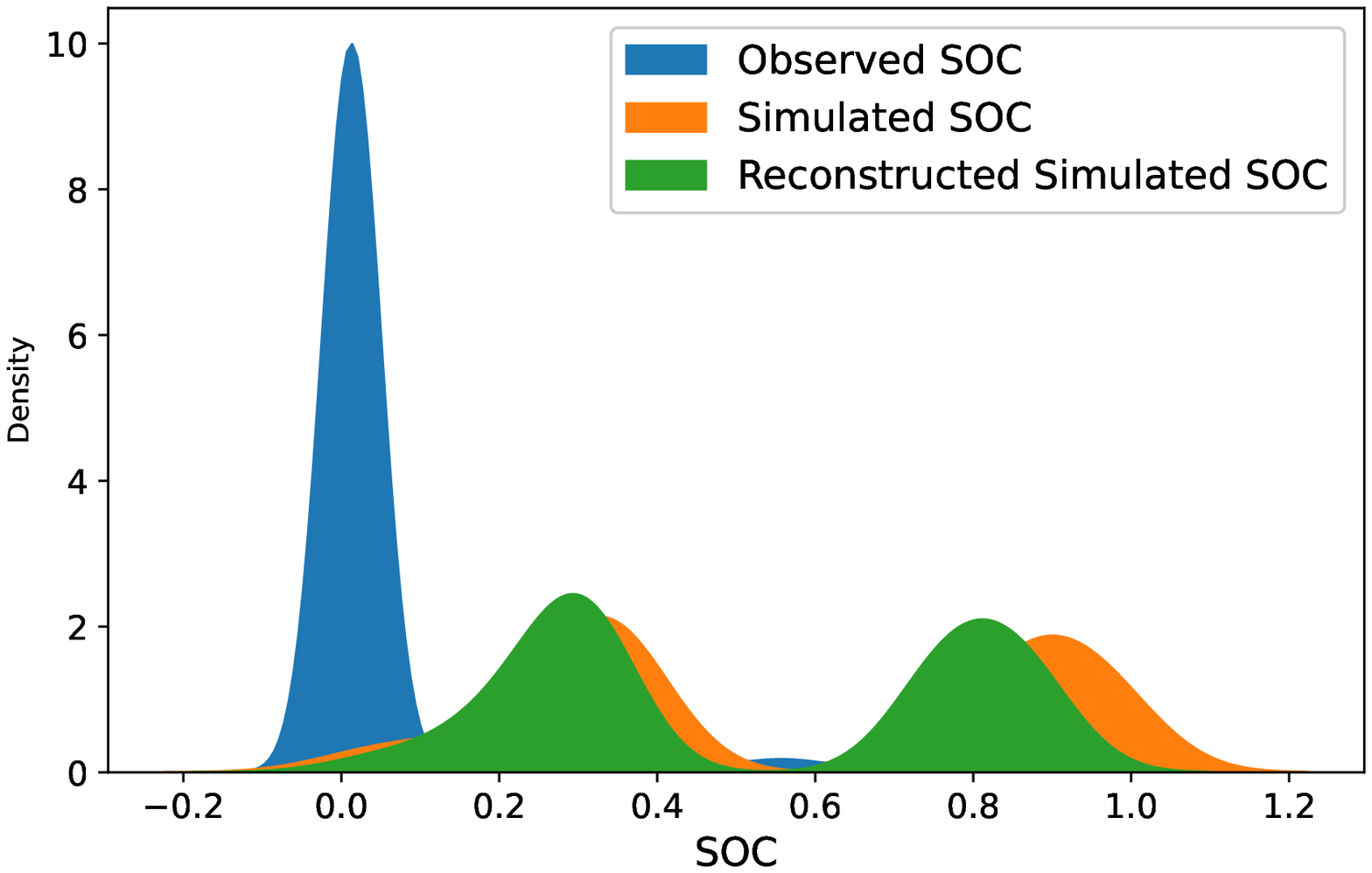}} \\
  \subcaptionbox{\label{fig:after-acdm-nw1} w DM, NW  \vspace{5pt} 
 }{\includegraphics[scale=0.16]{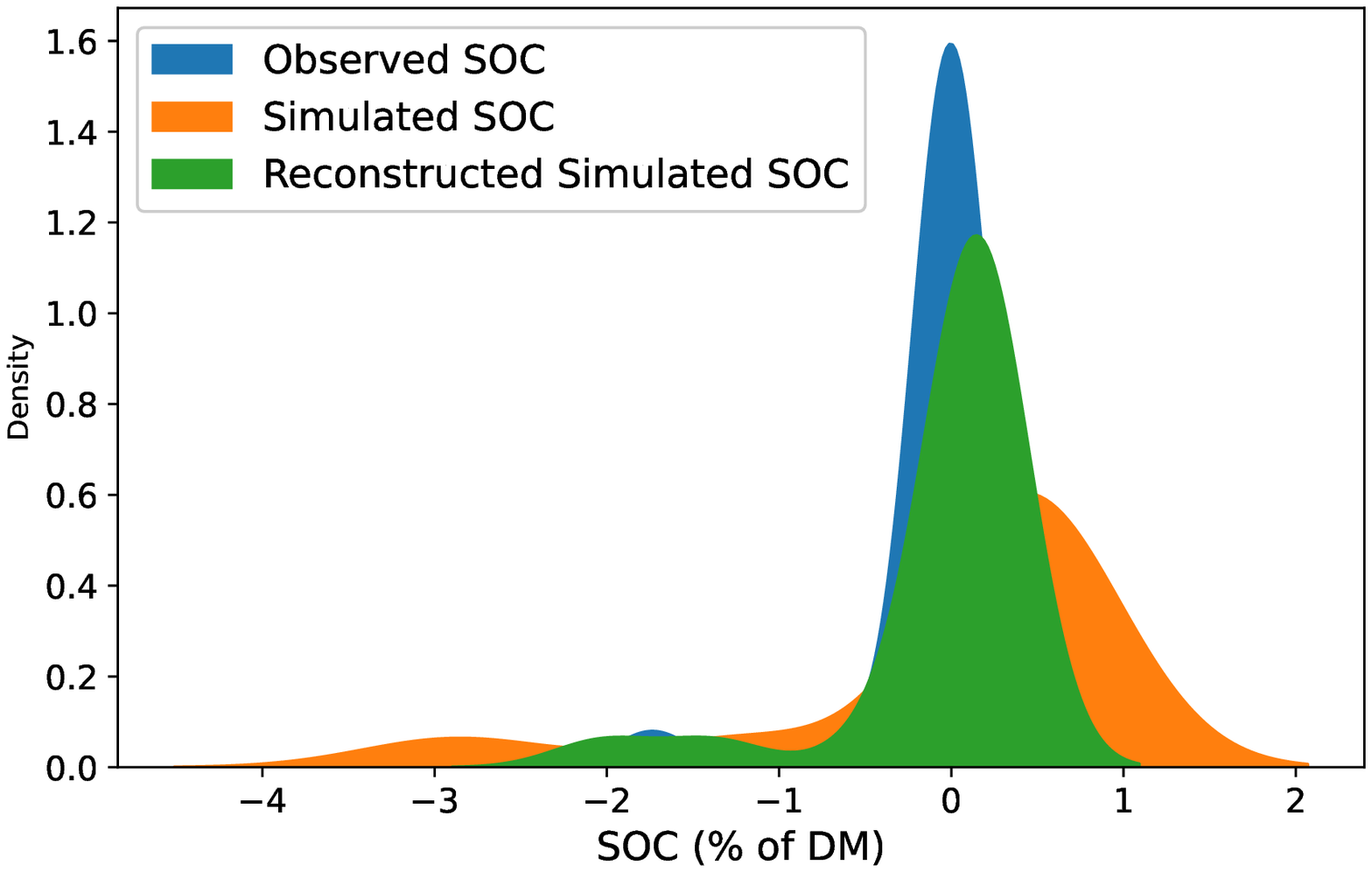}}
 &
 \subcaptionbox{\label{fig:after-acdm-gr} w DM, GR, NW  \vspace{5pt} }{\includegraphics[scale=0.16]{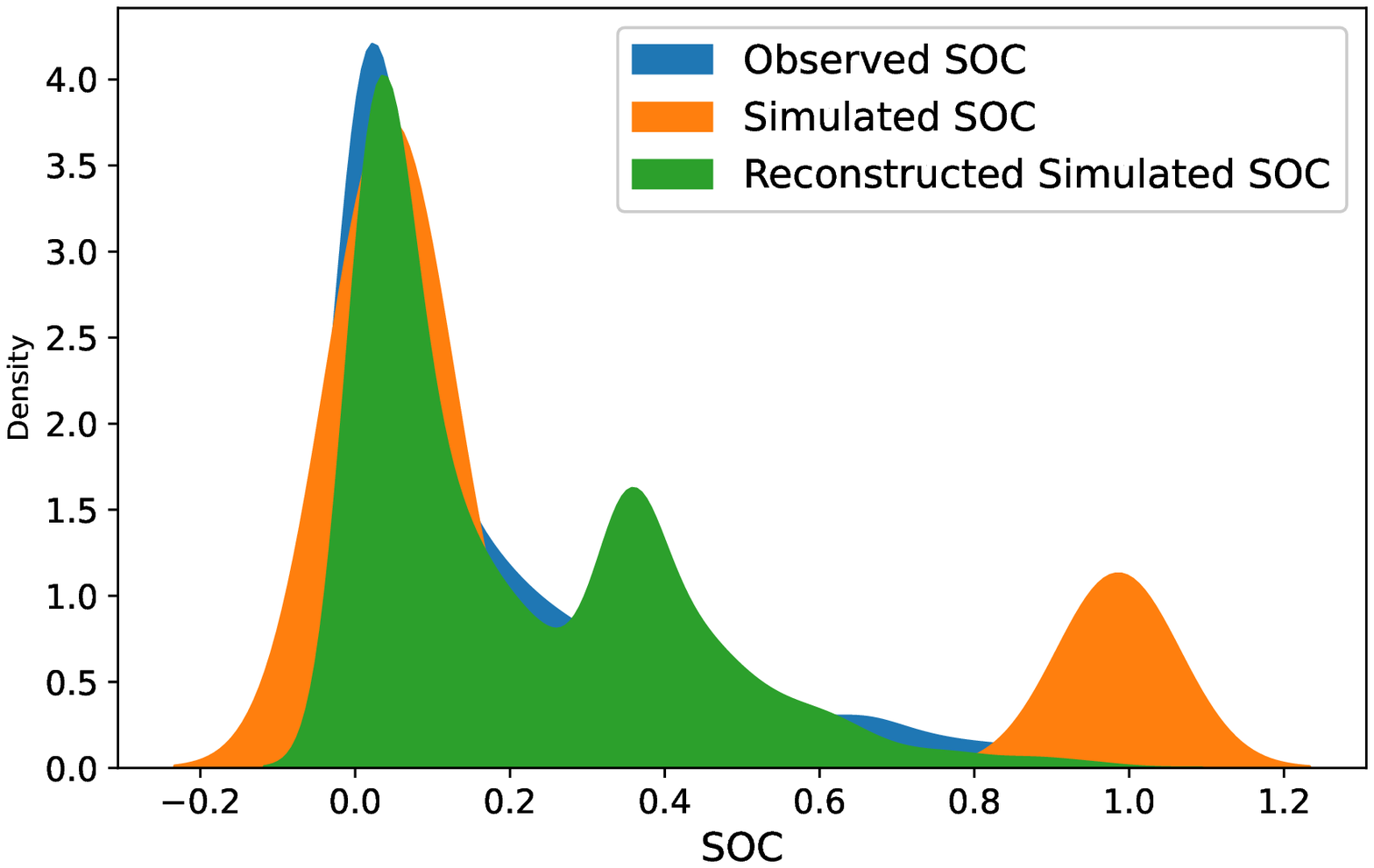}} &
 \subcaptionbox{\label{fig:after-acdm-gf} w DM, GF, NW  \vspace{5pt} }{\includegraphics[scale=0.16]{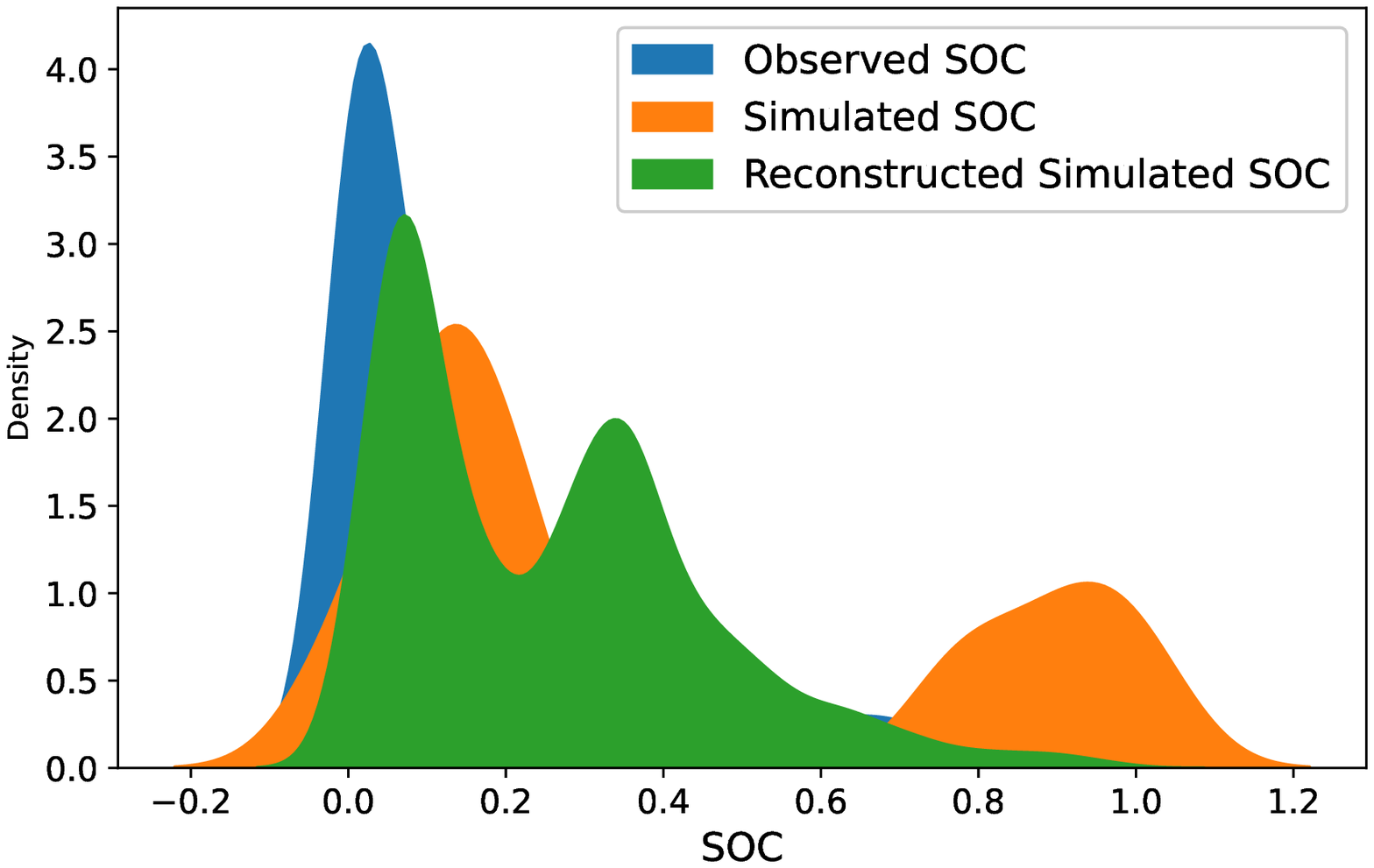}}&
 \subcaptionbox{\label{fig:after-acdm-shop} w DM, SHOP, WA  \vspace{5pt} }{\includegraphics[scale=0.16]{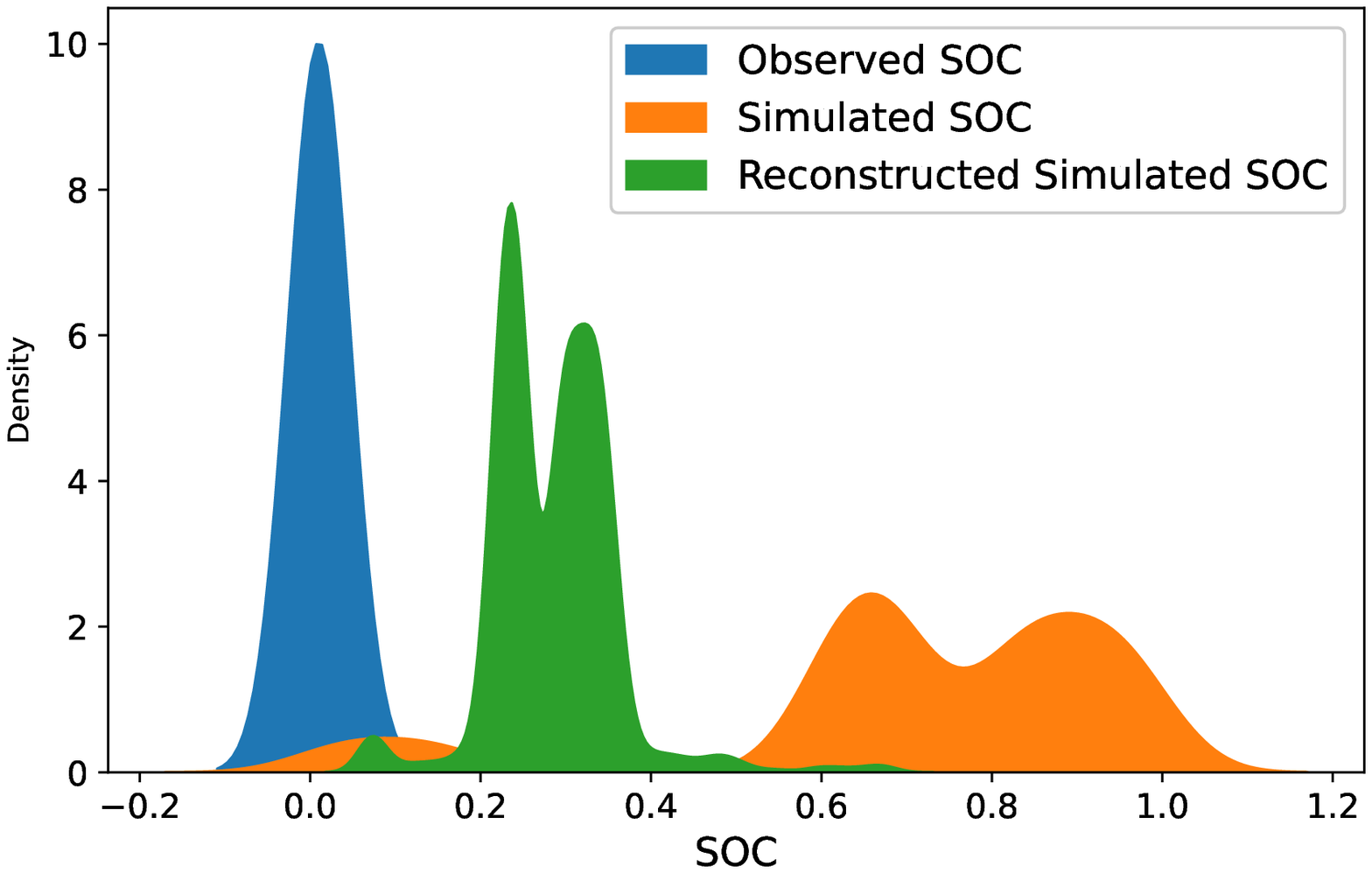}}&
 \subcaptionbox{\label{fig:after-acdm-hp} w DM, HP, WA  \vspace{5pt} }{\includegraphics[scale=0.16]{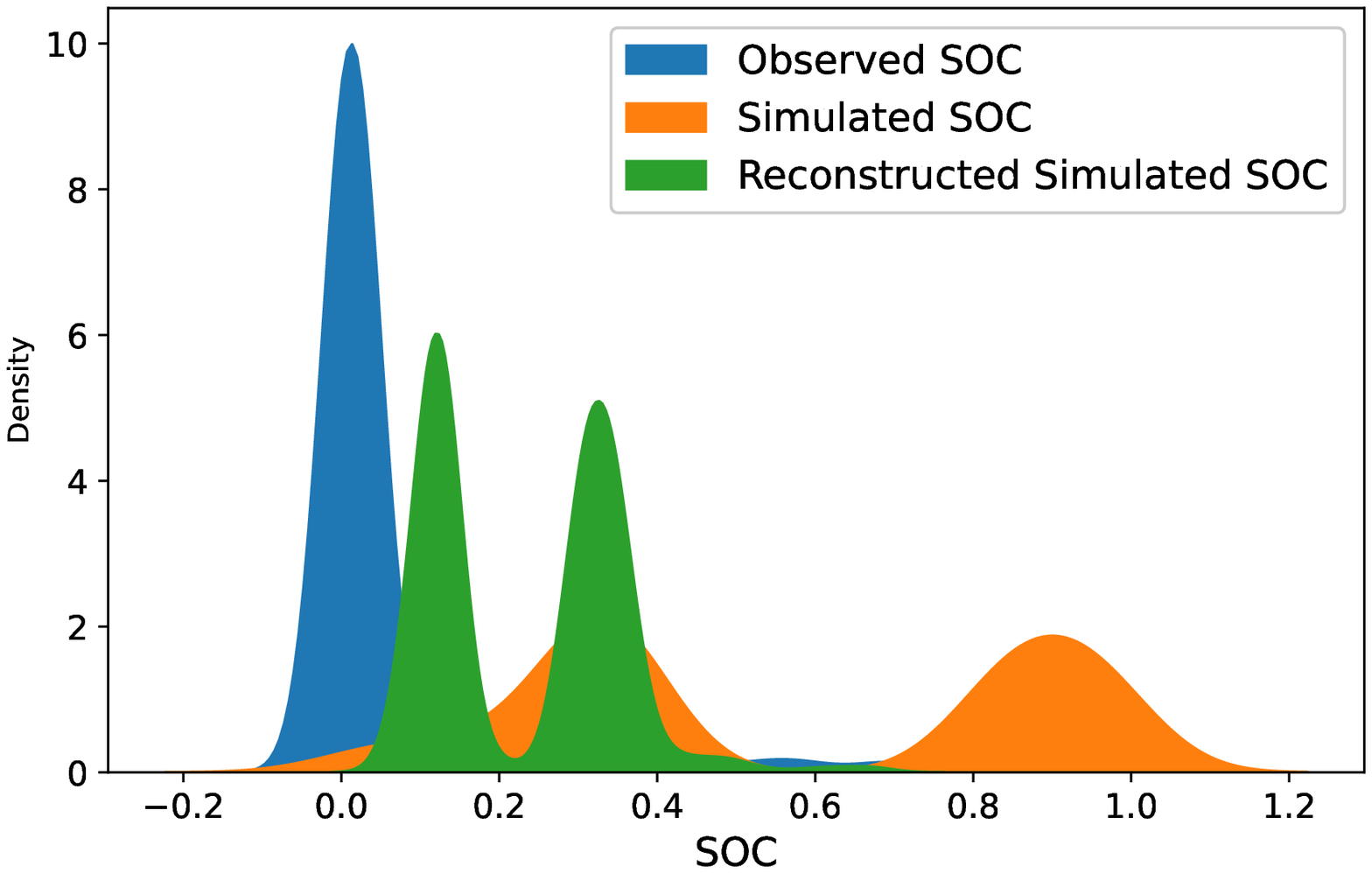}} \\ 
\end{tabular}
\caption{Effect of conditional distribution matching on simulated data. Soil carbon simulated data representations without (w/o) and with (w) distribution matching (DM) for North Wyke fields (NW), Golden Rove, NW (GR), Great Field, NW (GF), SHOP, WA and HomePlace, WA (HP). 
}
\label{fig:rl-plots-all}
\vspace{-10pt}
\end{figure*}





\begin{figure*}
    \centering
    \includegraphics[width=0.8\textwidth, height=0.8\textwidth]{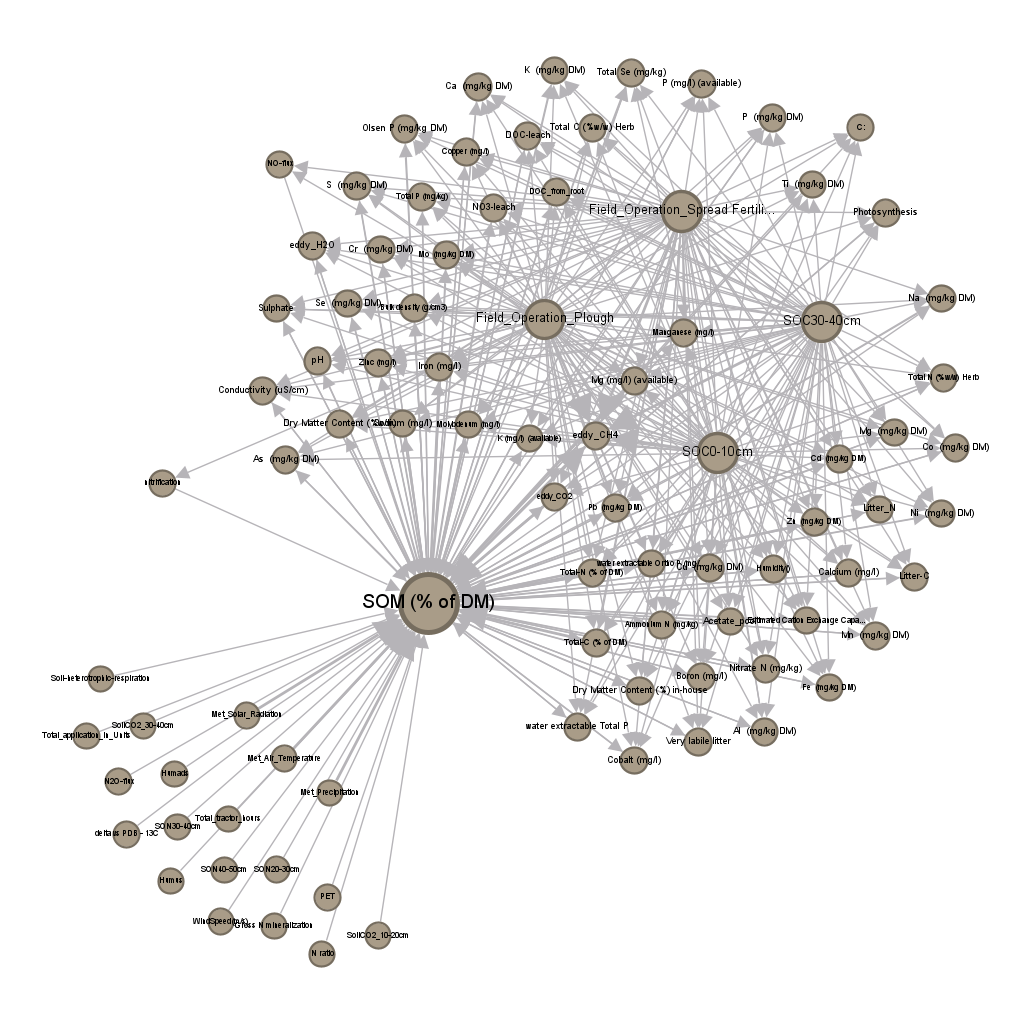}
    \caption{Discovered graph by KGRCL on North Wyke farm data.}
    \label{fig:kgrcl_nw_graph}
\end{figure*}

\begin{figure*}
    \centering
    \includegraphics[width=0.8\textwidth, height=0.8\textwidth]{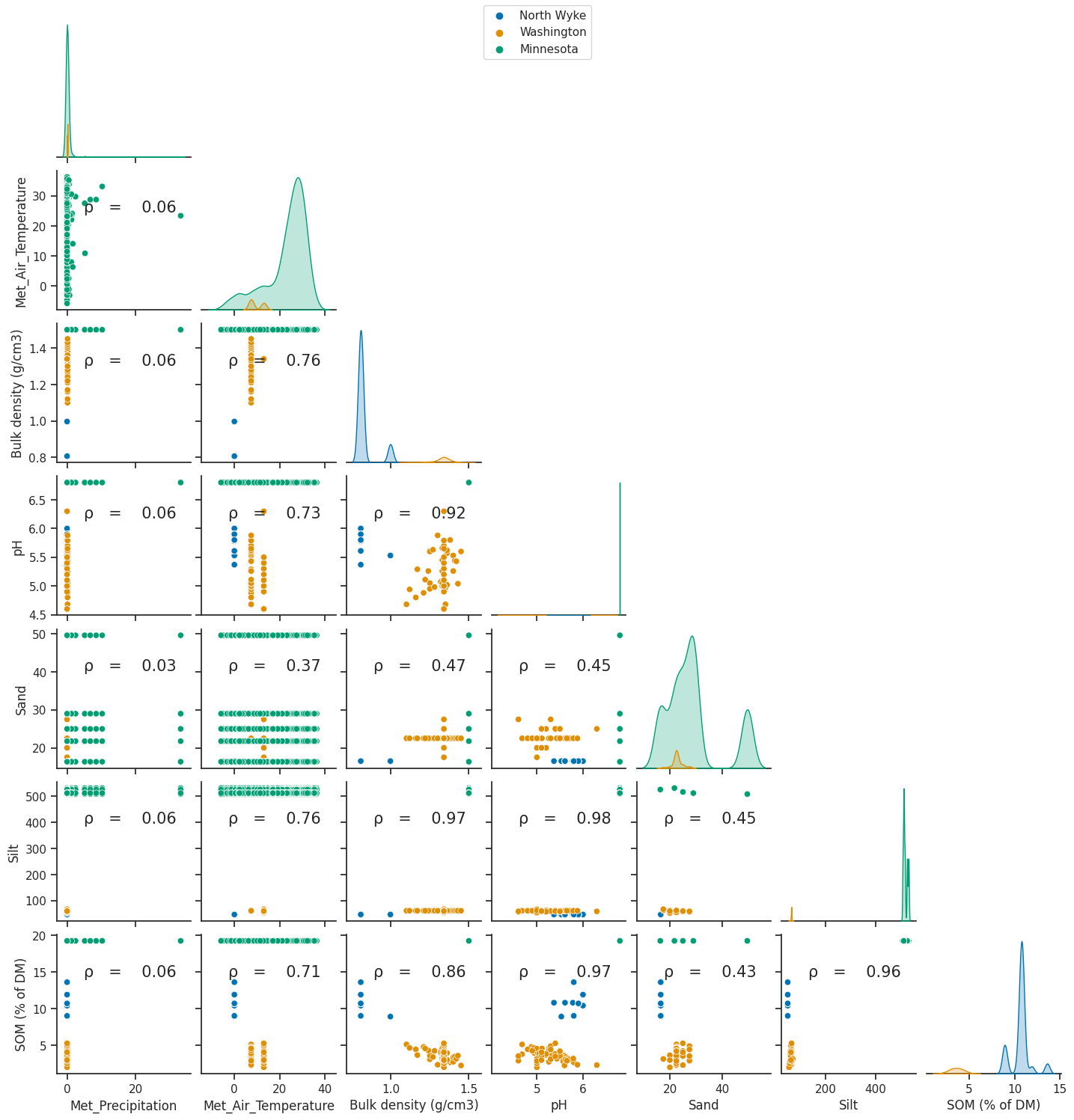}
    \caption{To aid researchers looking to reproduce our results on other soil data sets, we provide a detailed matrix plot to study the relationships between covariates across the three datasets, North Wyke, Washington and Minnesota. }
    \label{fig:pairplot}
\end{figure*}